\documentclass[graybox]{svmult}

\usepackage{type1cm}        
\usepackage{makeidx}         
\usepackage{graphicx}        
\usepackage{multicol}        
\usepackage[bottom]{footmisc}

\usepackage{newtxtext}       %
\usepackage[varvw]{newtxmath}       
\usepackage{xcolor}
\usepackage{verbatim} 
\usepackage{multirow}
\usepackage{booktabs}

\makeindex             

\newcommand{\xz}[1]{{\textcolor{cyan}{ #1} }}

\usepackage{threeparttable}
\usepackage[shortlabels]{enumitem}

\newcommand{\ttb}{\linespread{0.8}\bf\footnotesize\ttfamily}
\newcommand{\ttm}{\linespread{0.8}\footnotesize\ttfamily}

\usepackage{color}
\definecolor{ballblue}{rgb}{0.13, 0.67, 0.8}
\definecolor{bleudefrance}{rgb}{0.19, 0.55, 0.91}
\definecolor{altred}{rgb}{0.98,0.14,0.56}
\definecolor{deepred}{rgb}{0.7,0,0}
\definecolor{deepgreen}{rgb}{0,0.5,0}
\definecolor{lightgreen}{rgb}{0,0.6,0}

\usepackage{listings}

\let\origthelstnumber\thelstnumber
\makeatletter

\newcommand*\Suppressnumber{%
  \lst@AddToHook{OnNewLine}{%
    \let\thelstnumber\relax%
     \advance\c@lstnumber-\@ne\relax%
    }%
}

\newcommand*\Reactivatenumber[1]{%
  \setcounter{lstnumber}{\numexpr#1-1\relax}
  \lst@AddToHook{OnNewLine}{%
   \let\thelstnumber\origthelstnumber%
   \refstepcounter{lstnumber}
  }%
}

\newcommand\cppstyle{\lstset{
xleftmargin=3.0ex,
numbers=left,
numbersep=5pt,
escapeinside=||,
captionpos=b,
language=C++,
basicstyle=\ttm,
otherkeywords={self},             
keywordstyle=\ttb\color{bleudefrance},
emphstyle=\ttb\color{deepred},    
stringstyle=\color{deepgreen},
commentstyle=\color{lightgreen},
frame=tb,                         
showstringspaces=false            %
}}

\lstnewenvironment{cppenv}[1][]
{
\cppstyle
\lstset{#1}
}
{}

\newcommand\pythonstyle{\lstset{
xleftmargin=3.0ex,
numbers=left,
numbersep=5pt,
escapeinside=||,
captionpos=b,
language=Python,
basicstyle=\ttm,
otherkeywords={self},             
keywordstyle=\ttb\color{bleudefrance},
emphstyle=\ttb\color{deepred},    
stringstyle=\color{deepgreen},
commentstyle=\color{lightgreen},
frame=tb,                         
showstringspaces=false            %
}}

\lstnewenvironment{pythonenv}[1][]
{
\pythonstyle
\lstset{#1}
}
{}

\newcommand\pythoninline[1]{{\pythonstyle\lstinline!#1!}}

\begin{document}

\title*{Compilation and Optimizations for Efficient Machine Learning on Embedded Systems}
\author{Xiaofan Zhang, Yao Chen, Cong Hao, Sitao Huang, Yuhong Li, Deming Chen}
\institute{X. Zhang $\cdot$ Y. Li $\cdot$ D. Chen \at Electrical and Computer Engineering, University of Illinois at Urbana-Champaign, Champaign, IL, USA, \email{xiaofan3@illinois.edu; leeyh@illinois.edu; dchen@illinois.edu}
\and Y. Chen \at Advanced Digital Sciences Center, Singapore \email{yao.chen@adsc-create.edu.sg}
\and C. Hao \at Electrical and Computer Engineering, Georgia Institute of Technology, Atlanta, GA, USA \email{callie.hao@gatech.edu}
\and S. Huang \at Electrical Engineering and Computer Science, University of California Irvine, Irvine, CA, USA \email{sitaoh@uci.edu}
}
%
%
\maketitle

\abstract{
Deep Neural Networks (DNNs) have achieved great success in a variety of machine learning (ML) applications, delivering high-quality inferencing solutions in computer vision, natural language processing, and virtual reality, etc.
However, DNN-based ML applications also bring much increased computational and storage requirements, which are particularly challenging for embedded systems with limited compute/storage resources, tight power budgets, and small form factors. Challenges also come from the diverse application-specific requirements, including real-time responses, high-throughput performance, and reliable inference accuracy.
To address these challenges, we introduce a series of effective design methodologies, including efficient ML model designs, customized hardware accelerator designs, and hardware/software co-design strategies to enable efficient ML applications on embedded systems.
}


\section{Introduction}
\label{sec:intro}

The recent development of Deep Neural Networks (DNNs) has made machine learning based smart solutions more relevant and accessible to the general public. We have seen that some DNN technologies have been  integrated into our daily applications to provide high-quality inference services, such as image recognition, natural language processing, self-driving cars, and augmented and virtual reality \cite{lecun2015deep,he2016deep,vaswani2017attention,lombardi2018deep}, which have made our lives more convenient and our work more efficient. 
A significant number of these machine learning applications leverage edge devices and need to be deployed onto resource-constrained embedded systems, such as cell phones, cameras, and unmanned aerial vehicles (UAVs). They require not only higher inference accuracy to achieve intelligent responses but also aggressive inference speed, throughput, and energy efficiency to meet real-life demands.

As DNNs become more complicated, developing and serving the DNN-enabled applications requires more compute and memory resources, longer latency, and greater energy consumption. 
For example, the computation demands for DNN training have risen by over 300,000 times between AlexNet \cite{krizhevsky2012imagenet}, the champion model of the 2012 ImageNet competition, and the AlphaGo Zero \cite{silver2017mastering}, the AI player proposed in 2017 for the board game Go with superhuman skills \cite{openai_article}. By checking the image recognition models, there is a 16 times increase in model complexity from AlexNet with 85\% top-5 accuracy to ResNet-152 \cite{he2016deep} with 95\% top-5 accuracy. 

Such exponentially increasing compute and memory demands have created challenges and difficulties for DNN deployment on hardware, especially when targeting edge embedded devices with strictly limited compute and memory resources and tight power budgets \cite{zhao2012real, chen2016platform}. Although cloud computing can alleviate the burden of edge computing by taking over computationally intensive tasks, it is not always feasible when dealing with various real-life scenarios.
Primary reasons for sticking to edge embedded devices come from the unique requirements of the edge applications, which typically require real-time decision-making and reduced reliance on network communication and accessibility. 
They typically can not tolerate the extra latency caused by network data transfer due to the real-time response requirements.
In addition, private information, such as personal and sensitive data, should not be uploaded to the cloud without permission. It means that the edge devices are required to deliver not only high inference accuracy from DNNs, but also aggressive inference speed, throughput, and energy efficiency to meet various real-life demands.
In summary, the challenges of deploying machine learning workloads on edge embedded devices mainly come from three aspects:
1) DNN models are getting complicated and may fail to run efficiently, especially when targeting the low-power edge devices with scarce compute and memory resources;
2) Mapping DNN onto existing hardware or building domain-specific hardware is tedious and time-consuming;
3) Additional challenges come from inefficient optimization strategies that focus only on hardware or software optimizations alone but lack software/hardware co-design or cross-system stack design methods that can potentially deliver better overall solutions.

Despite the aforementioned challenges, there has been continuous progress in recent studies to explore various optimization strategies for edge machine learning solutions.
%
In this chapter, we present comprehensive design methodologies to face and overcome the challenges and enable efficient DNN applications on embedded systems. These methods include efficient DNN model designs in Sec. \ref{sec:eff_model}, accelerator design and workload mapping technologies in Sec. \ref{sec:compilers}, and cross-stack optimization strategies in Sec. \ref{sec:optimization}.



\section{Background and Related Works}
\label{sec:related_work}
Existing solutions to enable efficient DNN on embedded systems attempt to address challenges from the DNN model to the entire hardware-software system.
These different methods cover different development cycles and have different characteristics, as shown in Table~\ref{tab:methodattrib}.
In this section, we present existing work on the different design methods in terms of their different properties.

\begin{table}[]
\caption{Design methodologies and their attributes.}
\centering
\begin{tabular}{|l|l|}
\hline
Methods                                      & Attributes      \\ \hline
Efficient DNN model design                   & \begin{tabular}[c]{@{}l@{}}Design methods to create DNN models with less \\ parameters, less memory demands and less \\ computational complexity\end{tabular} \\ \hline
Efficient accelerator design and DNN mapping & \begin{tabular}[c]{@{}l@{}}Solutions to build domain specific \\ hardware/software accelerators with optimized \\ task scheduling\end{tabular}  \\ \hline
Efficient DNN/accelerator co-design          & \begin{tabular}[c]{@{}l@{}}Optimization strategies that integrate both the \\ hardware design process and DNN algorithm \\ design process\end{tabular}  \\ \hline
\end{tabular}%
\label{tab:methodattrib}
\end{table}

\subsection{Efficient DNN designs}

A DNN includes multiple intermediate layers between the input and output layers, and each intermediate layer consists of artificial neurons for transforming the input information (e.g., input feature maps) following the predefined network connection.
In general, a DNN contains millions of parameters and requires billions of operations during inference. To successfully deploy DNNs onto hardware with desired performance, developers focus on network compression to reduce network complexities and lower the compute and memory demands. Recent research has demonstrated the possibility of using quantized data to represent original floating-point parameters, such as using 8-bit quantization or even binary and ternary data representation \cite{jouppi2017datacenter, rastegari2016xnor,wang2018design,gope2019ternary,gong2020vecq,chen2019tdla}. These solutions are intended to replace the hardware-intensive floating-point multiplications by logical operations so that DNNs can be more efficient on hardware platforms.

Another method to compress DNN is network pruning, which aims to reduce the redundancy of DNN structures \cite{han2015learning,han2016deep,luo2017thinet}. According to the published pruning strategies, the less essential connections between DNN layers are discarded, and network retraining is then performed to regain accuracy. Significant reductions can be achieved on the classic DNNs, such as AlexNet \cite{krizhevsky2012imagenet} and VGG-16 \cite{simonyan2014very}. Since the major benefit of network compression comes from the fully-connected (FC) layers, to continuously have effective pruning results for latter DNNs (e.g., GoogleNet \cite{szegedy2015going} and ResNet \cite{he2016deep}) with fewer FC layers, more sophisticated algorithms are required to achieve effective network pruning, such as using evolutionary algorithms \cite{dai2019nest}, alternating direction method of multipliers \cite{ren2019admm}, and iterative pruning \cite{ding2018auto}.

As most of the computations happen inside the convolutional (Conv) layers, previous works also attempt to reduce the computational complexity by using depth-wise separable Conv layers \cite{howard2017mobilenets}. The depth-wise separable structure can effectively reduce the number of operations and provide more compact DNN designs for resource-constrained hardware. To further improve the DNN deployment on hardware, layer fusion is proposed in \cite{alwani2016fused} to minimize data movements between on-chip and off-chip memory.


\subsection{Efficient accelerator designs and DNN mapping methods}
Building domain-specific hardware accelerators is another popular approach for efficient DNN deployment. These accelerators attempt to take advantage of customized or specialized hardware and software designs, such as adopting acceleration libraries on CPUs \cite{intel_article}, exploring kernel optimization on GPUs \cite{franklin2018nvidia}, and building customized accelerators on FPGAs \cite{zhang2015optimizing,qiu2016going,zhang2018dnnbuilder} and ASICs \cite{isscc_2016_chen_eyeriss, han2016eie,jouppi2017datacenter} to improve the speed and efficiency of DNN inference and training processes.
Among these accelerator designs, FPGA- and ASIC-based designs can be fully customized to implement the neural network functionality with improved latency, throughput, and energy consumption compared to CPU- and GPU-based designs. 

Still, developing customized accelerators present significant challenges, such as the tedious hardware design process, the intricate hardware verification problems, and the time-consuming design space exploration during DNN deployment. 
To alleviate these challenges, recent investigations have started focusing on techniques including high-level synthesis \cite{papakonstantinou2009fcuda,rupnow2011study,liu2016high} and end-to-end design frameworks for fast DNN accelerator design and efficient workload deployment \cite{zhang2018dnnbuilder, zhang2018caffeine,ye2020hybrid,huang2021pylog}. 
%
They support high abstraction inputs, such as Python-based DNN descriptions used by popular machine learning frameworks (e.g., Caffe  \cite{jia2014caffe}, TensorFlow \cite{abadi2016tensorflow}, PyTorch \cite{paszke2019pytorch}), so DNNs can be directly imported without manual code conversions and be parsed and then mapped onto hardware. 
These frameworks, such as DNNBuilder \cite{zhang2018dnnbuilder} and HybridDNN \cite{ye2020hybrid} also integrate design space exploration (DSE) engines to perform effective and systematical explorations and deliver highly optimized accelerators to meet the user-specific requirements.

\subsection{Efficient co-design optimization}

Recent research also focuses on cross-stack co-design optimizations to enable successful DNN deployment on embedded systems~\cite{hao2021enabling}.
%
Instead of independently optimizing hardware and software components, researchers proposed algorithm/accelerator co-design and co-search to solve the edge AI challenges: DNNs are designed to satisfy accuracy demands and must be aware of the hardware constraints with rational network configurations. At the same time, the accelerators need to provide extensive support for different DNN components without introducing too many restrictions on network design and guarantee performance to meet the specifications. 
The authors in \cite{hao2018deep} proposed the concept of DNN/accelerator co-design for the first time, which aims to consider software and hardware metrics simultaneously to ``automatically generate both DNN models and their corresponding implementations as pairs". This concept is then demonstrated by winning the competitive System Design Contest for low power object detection in the 56th IEEE/ACM Design Automation Conference (DAC-SDC) \cite{zhangskynet}.

Many follow-up works continued investigating the co-design opportunities between different AI algorithms and hardware devices~\cite{hao2019fpga,yang2019synetgy,guo2019dl,hao2019nais,jiang2020hardware,wang2021exploring,zhang2021exploring,fu2021a3c}.
These co-design approaches have been studied with remarkable achievements by combining multiple optimization techniques across both hardware and software.
For example, while neural architecture search (NAS) has been largely successful in designing high-quality DNN models~\cite{elsken2019neural,zoph2018learning},
hardware-aware NAS is drawing increasing attention, which aims at delivering high-accuracy models with hardware efficiency as well (e.g., FBNet~\cite{wu2019fbnet} and MNasNet~\cite{tan2019mnasnet}).
Other machine-learning algorithm/hardware co-design works include FNAS~\cite{jiang2019accuracy}, NAIS~\cite{hao2019nais}, EDD~\cite{li2020edd}, and NASAIC~\cite{yang2020co}.
Driven by the success of such a co-design strategy, other types of co-design methods are also proposed recently, including software/compiler co-design~\cite{ma2020pconv, niu2020patdnn, ji2020mcunet}, compiler/hardware co-design~\cite{pulp-main, garofalo2020xpulpnn, garofalo2020pulp}, etc.

\section{Efficient Machine Learning Model Designs}
\label{sec:eff_model}


Machine learning applications require not only high inference accuracy but also aggressive inference speed, throughput, and energy efficiency to meet real-life demands. They rely on hardware-efficient DNN designs, especially when targeting edge scenarios with limited hardware resources.
In this section, we introduce ELB-NN \cite{wang2018design} and VecQ \cite{gong2020vecq} to deliver hardware-efficient DNNs for embedded systems.

\subsection{The ELB-NN}

ELB-NN (Extremely Low Bit-width Neural Network) is proposed to enhance energy efficiency when running image classification on an embedded FPGA. It is one of the first hybrid low-bit-width designs that supports arbitrary DNN quantization.
This subsection presents the hybrid quantization feature of the ELB-NN and its corresponding hardware accelerator design on embedded systems.

\subsubsection{Hybrid quantization scheme}

Hybrid quantization means that different quantization schemes are involved for the network's parameters and activations. The quantization scheme can go all the way down to binary. To better adapt the hybrid quantization, we first investigate their impacts on the network inference accuracy.
We follow Eq. \ref{eq:elb_nn_binary} to calculate the binary weights. Here $\tilde{w}$ represents the full precision weights after back propagation, while {\bf $E(|\tilde{w}|)$} represents the mean of all the full-precision weights as a scaling factor. For the ternary training, the $w_t$ (representing ternary parameters) can be calculated following Eq. \ref{eq:elb_nn_ternary}. Here we set the threshold $w_{thres}=0.7E(|\tilde{w}|)$ and calculate the scaling factor $E$ as suggested in \cite{li2016ternary}. 
We also apply relatively high precision using 8- or 4-bit fixed-point representation.
We then use AlexNet \cite{krizhevsky2012imagenet} to perform quantitative analysis when applying hybrid quantization.

\begin{equation}
w_b = sign(|\tilde{w}|)\times\textbf{E}(|\tilde{w}|)
\label{eq:elb_nn_binary}
\end{equation} 
\begin{equation}
w_t = 
\begin{cases}
sign(\tilde{w})\times $E$ & \mbox{$|w_t|>w_{thres}$}\\
0 & \mbox{$|w_t| \leq w_{thres}$}
\end{cases}
\label{eq:elb_nn_ternary}
\end{equation}

\begin{figure}
    \begin{centering}
    \includegraphics[width=0.75\columnwidth]{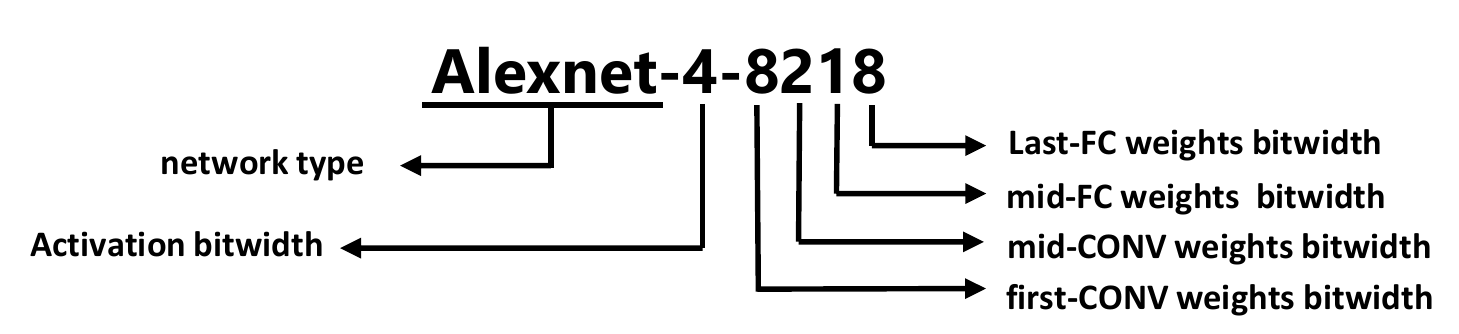}
    \caption{Network representation when using hybrid quantization \cite{wang2018design}}    
    \vspace{+2pt}
    \label{fig:elb_nn_representation}
    \end{centering}
\end{figure}
 
In this analysis with AlexNet, we focus on the impact of 1) the quantized parameters of the convolutional (CONV) and fully-connected (FC) layer, and 2) the quantized activations. 
We use mid-CONV to denote all the CONV layers except the first CONV layer and mid-FC to denote all the FC layers except the last FC layer. The naming rule of the proposed hybrid precision network can be referred to Fig. \ref{fig:elb_nn_representation}. 

\begin{table}
  \centering
  \caption{Inference accuracy with hybrid quantization using ImageNet dataset \cite{wang2018design}.}
  \label{tab:elb_nn_accuracy}
  \begin{tabular}{|c|c|}
  \hline
    Network precision & Accuracy (Top-1)\\
    \hline
    Alexnet with float32 & 55.9\%\cite{zhou2016dorefa} \\
    \hline
    Alexnet-8-8888 & 54.6\% \\
    \hline
    Alexnet-8-8228 & 53.3\% \\
     \hline
    Alexnet-8-8218 & 52.6\% \\
    \hline
    Alexnet-8-8118 & 51.1\% \\
    \hline
    Alexnet-4-8218 & 49.3\% \\
    \hline
    Alexnet-2-8218 & 46.1\% \\
    \hline
    Alexnet-4-8218 (w/o g.) & 53.2\%\\
    \hline
    Alexnet-4-8218 (ext.) & 54.5\%\\
    \hline
  \end{tabular}
\end{table}

In Table \ref{tab:elb_nn_accuracy}, the 8-bit design (Alexnet-8-8888) only reduces the accuracy by 1.3\% compared to the original float32 version. The accuracy is still promising after using ternary (Alexnet-8-8228) and binary (Alexnet-8-8218) parameters for mid-CONV and mid-FC layer. It means that the network is relatively robust to the precision of parameters. On the contrary, the precision of activations significantly impacts classification accuracy. Compared to the Alexnet-8-8218, we observe 3.3\% and 6.5\% accuracy drop when activations move to 4 bits (Alexnet-4-8218) and 2 bits (Alexnet-2-8218). 
To further investigate, we disable the group function, which was originally proposed to handle the limited GPU memory issue. As a result, we capture an 80\% computation increase and a 3.9\% accuracy improvement in Alexnet-4-8218 (w/o g.). We further increase the channel number for the five CONV layers 
in Alexnet-4-8218 (ext.)
and achieve 1.3\% accuracy gain by affording extra 61\% computation compared to Alexnet-4-8218 (w/o g.). By increasing the model complexity, we can bring back the accuracy.
These observations are insightful for hybrid DNN quantization as parameters can be quantized more aggressively than activations. 

\subsubsection{Hardware accelerator for ELB-NN}
\label{sec:elb_nn_accelerator}
\begin{figure}[h]
    \begin{centering}
    \includegraphics[width=0.8\columnwidth]{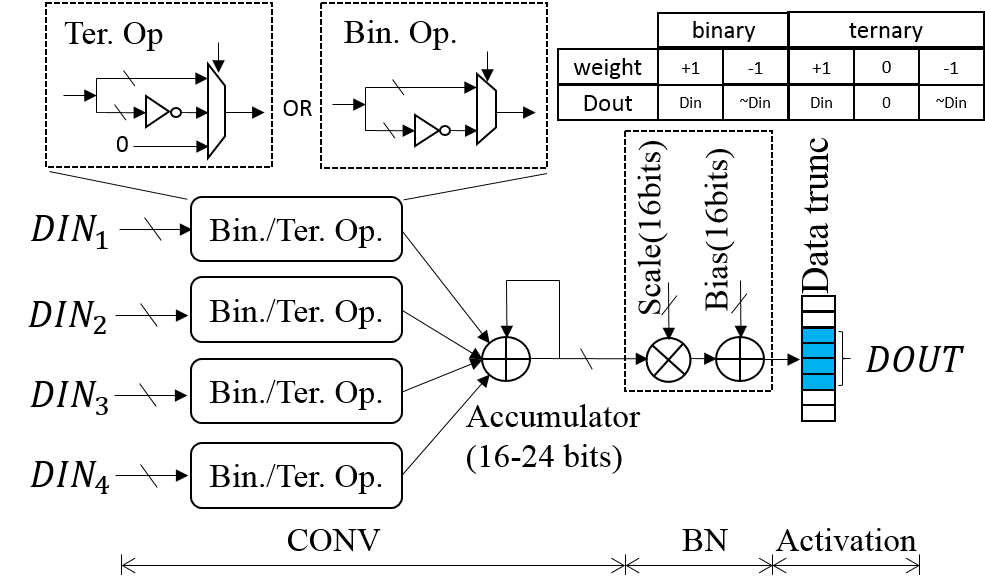}
    \caption{Computation engine (CE) with binary and ternary logic operations \cite{wang2018design}.}
    \label{fig:elb_nn_kernel}
    \end{centering}
 \end{figure}

To handle ELB-NN, we propose a parameterized Computation Engine (CE) in \cite{wang2018design} with flexible support of low bit-width quantization and configurable parallelism during execution.
As shown in Fig. \ref{fig:elb_nn_kernel}, it contains a four-input-parallel CE as an example, where four inputs can be processed simultaneously (including binary/ternary operations and accumulation) and then followed by batch normalization (BN) and activation function. 
The precision of the accumulator is adjustable, which is intended to allow more flexible quantization designs and maintain the output accuracy. For a larger number of inputs, an adder tree will be used before the accumulator for timing enclosure. 

To demonstrate the hardware efficiency of ELB-NN, we adopt the accelerator with the proposed CE and accelerate different quantized versions of the AlexNet and VGG16 using an embedded platform called ZC706 (with an ARM CPU and a Xilinx Kintex-7 FPGA).
Results are shown in Table \ref{tab:eln_nn_performance}. ELB-NN can achieve throughput performance up to 10.3 TOPS, which outperforms previous designs in \cite{zhao2017accelerating, umuroglu2017finn, Nurvitadhi2016AcceleratingBN}.

\begin{table*}[h]
\scriptsize
\vspace{-2pt}
\caption{ELB-NN performance evaluated on an embedded platform (Xilinx ZC706) \cite{wang2018design}.}
\label{tab:eln_nn_performance}
\vspace{-6pt}
\begin{center}
\newcommand{\tabincell}[2]{\begin{tabular}{@{}#1@{}}#2\end{tabular}}
\begin{tabular}{|c|c|c|c|c|c|c|c|c|c|}
\hline
\multirow{2}{*}{Network} & \multicolumn{4}{c|} {Utilization}  & \multirow{2}{*}{\tabincell{c}{Batch\\size}} & \multirow{2}{*}{\tabincell{c}{Bandwidth\\(GBytes/s)}} & \multirow{2}{*}{\tabincell{c}{Complexity\\(GOP)}} & \multirow{2}{*}{\tabincell{c}{imges/s}} & \multirow{2}{*}{\tabincell{c}{Perf.\\(TOPS)}}\\ 
\cline{2-5} 
 & LUT  & FF  & BRAM & DSP  & \multicolumn{1}{c|}{} & \multicolumn{1}{c|}{} & \multicolumn{1}{c|}{} & \multicolumn{1}{c|}{} & \multicolumn{1}{c|}{} \\
\hline 
Alexnet-8-8888 & \tabincell{c}{86262(39\%)} & \tabincell{c}{51387(12\%)} & \tabincell{c}{303(56\%)} & \tabincell{c}{808(90\%)} & 2 & 10.8 & 1.45 & 340 & 0.493 \\ 
\hline
Alexnet-8-8218 & \tabincell{c}{103505(47\%)}  & \tabincell{c}{90125(21\%)} & \tabincell{c}{498(91\%)} & \tabincell{c}{550(61\%)} & 5 & 3.35 & 1.45 & 856.1 & 1.24 \\
\hline
Alexnet-4-8218 & \tabincell{c}{105673(48\%)} & \tabincell{c}{94149(22\%)} & \tabincell{c}{463(85\%)} & \tabincell{c}{880(98\%)} & 8 & 3.35 & 1.45 & 1369.6 & 1.99 \\ 
\hline
Alexnet-4-8218 (w/o g.) & \tabincell{c}{127393(58\%)}  & \tabincell{c}{105328(24\%)} & \tabincell{c}{435(80\%)} & \tabincell{c}{839(93\%)} & 7 & 4.30 & 2.61 & 1198.5 & 2.59 \\
\hline
Alexnet-4-8218 (ext.) & \tabincell{c}{124317(57\%)}  & \tabincell{c}{101558(23\%)} & \tabincell{c}{481(88\%)} & \tabincell{c}{783(87\%)} & 7 & 3.4 & 4.22 & 599.2 & 2.53 \\
\hline
\hline
VGG16-4-8218 & \tabincell{c}{112992(52\%)}  & \tabincell{c}{99396(23\%)} & \tabincell{c}{509(93\%)} & \tabincell{c}{298(33\%)} & 2 & 5.85 & 31.0
 & 110.7 & 3.43 \\
\hline
VGG16-2-8118 & \tabincell{c}{137973(63\%)} & \tabincell{c}{113262(26\%)} & \tabincell{c}{499(92\%)} & \tabincell{c}{651(72\%)} & 3 & 6.67 & 31.0
 & 332.2 & 10.3 \\
\hline
\end{tabular}
\end{center}
\end{table*}

\subsection{The VecQ}

Vectorized Quantization (VecQ) is a training quantization framework that is based on a novel quantization loss measurement metric called vector loss. It is proposed to provide flexible bitwidth support with minimal quantization loss to achieve higher model accuracy.
In this subsection, we present the detailed definition of vector loss and the VecQ quantization framework.

\subsubsection{Quantization with Vector Loss}

We use the square of the Euclidean distance of the data before and after quantization to represent quantization loss. It is also called Square 2-norm or L2 distance.
Minimizing the L2 loss during the model training is proved to be adequate in providing higher model accuracy~\cite{hubara2016binarized,Binaryconnect,Ternaryconnect,zhou2016dorefa,jin2018sparse}.
However, due to the non-convex characteristic of optimization for L2 loss under a constrained bitwidth requirement, the quantization easily falls into sub-optimal solution space. In addition, adopting the L2 distance collects the loss of each quantized data individually and neglects the distribution and correlations among these data points in a kernel or a layer.

Focusing on the limitations above, in VecQ, we first flatten and reshape the weight set $\mathrm{W_f}(l)$ for a layer of the DNN and reshape them as a vector $\boldsymbol{w_f}(l)$ with the dimension of the size of the elements. For example, it will be a $N \times M \times K^2$-dimensional vector for a CNN layer with $N$ input channel, $M$ output channel and $K$ size of filter kernel. The quantized weight vector is denoted as $\boldsymbol{w_q}(l)$.

\begin{figure}
    \centering
    \includegraphics[width=0.4\columnwidth]{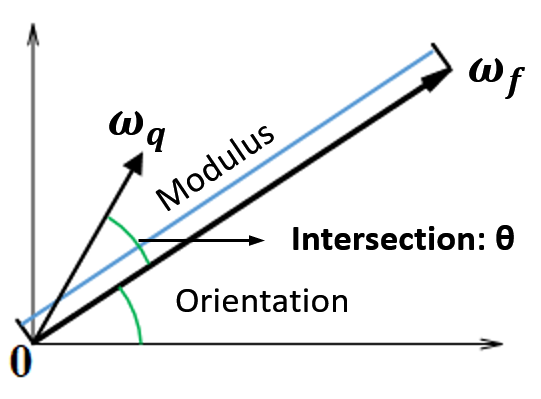}
    \caption{The attributes of a vector and the quantization angle of $w_f$ and $w_q$.}
    \label{fig:vectorloss}
\end{figure}

There are two attributes of a vector, as shown in Fig.~\ref{fig:vectorloss} orientation and modulus. We define a quantization angle representing the intersection angle between the original weight vector and the quantized vector. So, the vector distance between the weight vector before and after quantization is determined by the quantization angle and the vector's modulus.
We then define a vector loss, denoted as $J_v$, and compose it with the orientation loss $J_o$ and the modulus loss $J_m$.
\begin{equation}
 J_v = J(\boldsymbol w_{f}, \boldsymbol w_{q}) = J_o(\boldsymbol w_{f}, \boldsymbol w_{q}) + J_m(\boldsymbol w_{f}, \boldsymbol w_{q})
\end{equation}

Where $J_o$ and $J_m$ are computed as:
\begin{equation}
\begin{split}
    J_o&=1-\cos\theta,~(\cos\theta=\frac{\alpha\boldsymbol w_q}{|\alpha\boldsymbol w_q|}\frac{\boldsymbol w_f}{|\boldsymbol w_f|}) \\
    &=1-\boldsymbol{e_{v}} \boldsymbol{e_{w_f}}\\
    &=1-\sum_{i=1}^{d}(e_{v_i}e_{w_{fi}}) \\
    J_m&=||\boldsymbol w_f-\alpha\boldsymbol w_q||_2^2
\end{split}
\label{eq:decomposition}
\end{equation}
here, the $\boldsymbol{e_v}$ and $\boldsymbol{e_{w_f}}$ represent the unit vector for $\boldsymbol{v}$ and $\boldsymbol{w_f}$.
$\boldsymbol{w_f}$ is a weight vector of a layer of a DNN containing $d$ weights.

\begin{figure}[h]
    \centering
    \includegraphics[width=0.7\textwidth]{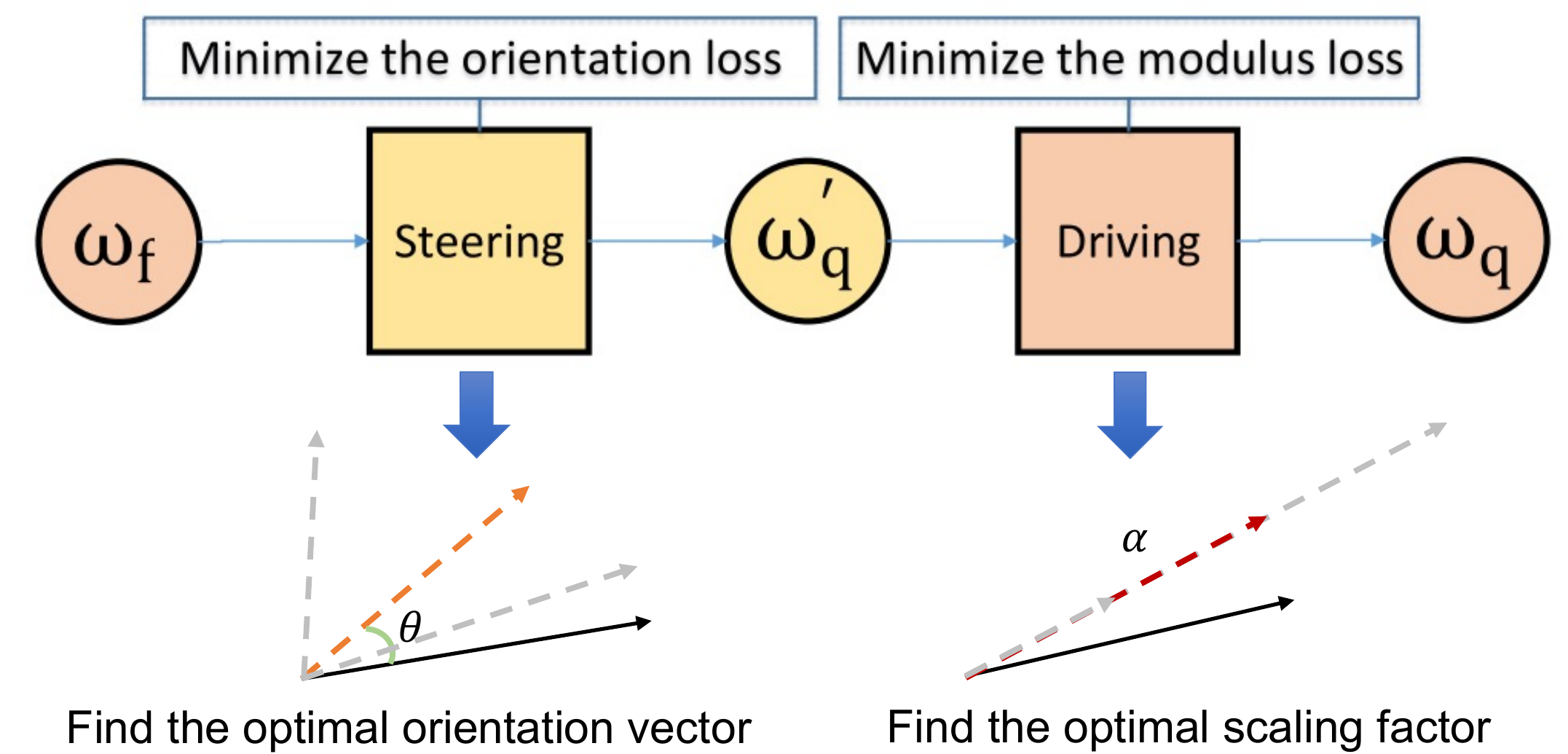}
    \caption{The overall flow of quantization process, including both steering and driving stage~\cite{gong2020vecq}.}
    \label{fig:quantization_process}
\end{figure}

With these approaches, the orientation loss $J_{o}$ indicates the optimized quantization angle and the modulus loss $J_{m}$ indicates the optimized scale at this angle.
Therefore, our quantization takes two stages to minimize the two losses independently,
which are defined as \textbf{steering stage} and \textbf{driving stage} as shown in Fig.~\ref{fig:quantization_process}. 
In the steering stage, we adjust the orientation of the weight vector to minimize the orientation loss.
Then, we fix the orientation and only scale the modulus of the vector at the driving stage to minimize the modulus loss.

\subsubsection{Framework integration}

\begin{figure*}
    \centering
    \includegraphics[width=0.99\textwidth]{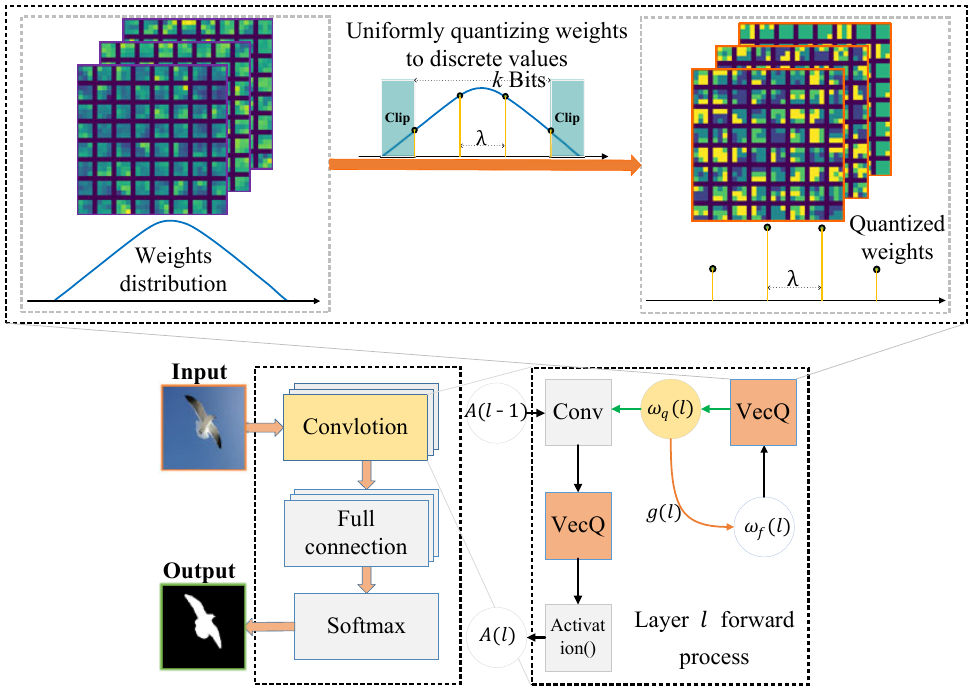}
    \caption{Integrated quantization process in DNN training~\cite{gong2020vecq}.}
    \label{fig:qinte}
\end{figure*}

The VecQ quantization is integrated into the DNN training flow for both the weight data and the activation data. As shown in Fig.~\ref{fig:qinte}.
For weight data, taking a layer $l$ as an example, during the forward propagation, the weights $w_f(l)$ represented in floating-point is quantized into $w_q(l)$, then use the quantized weights to compute the output of this layer. 
To simplify the computing process, the weight is treated as normally distributed and an interval $\lambda$ is used regarding the given bitwidth constraint.
During the backward propagation, the gradient is calculated with $w_q(l)$ instead of $w_f(l)$ and propagated. In the final update process, the gradient $g(l)$ of $w_q(l)$ is updated to $w_f(l)$~\cite{zhou2016dorefa}.

For the activation output of a layer, during the training, we compute a distribution parameter of the activation outputs $p(t)$ and update it with Exponential Moving Average. 
During the inference, the distribution parameter is employed as a linear factor to the activation function~\cite{ioffe2015batch}.
The $A(l)$ is the activation output of layer $l$, and $Activation(\cdot)$ is the non-linear activation function following the convolution or fully-connected layers, such as Sigmoid, Tanh, ReLU.

\begin{figure}
    \centering
    \includegraphics[width=\textwidth]{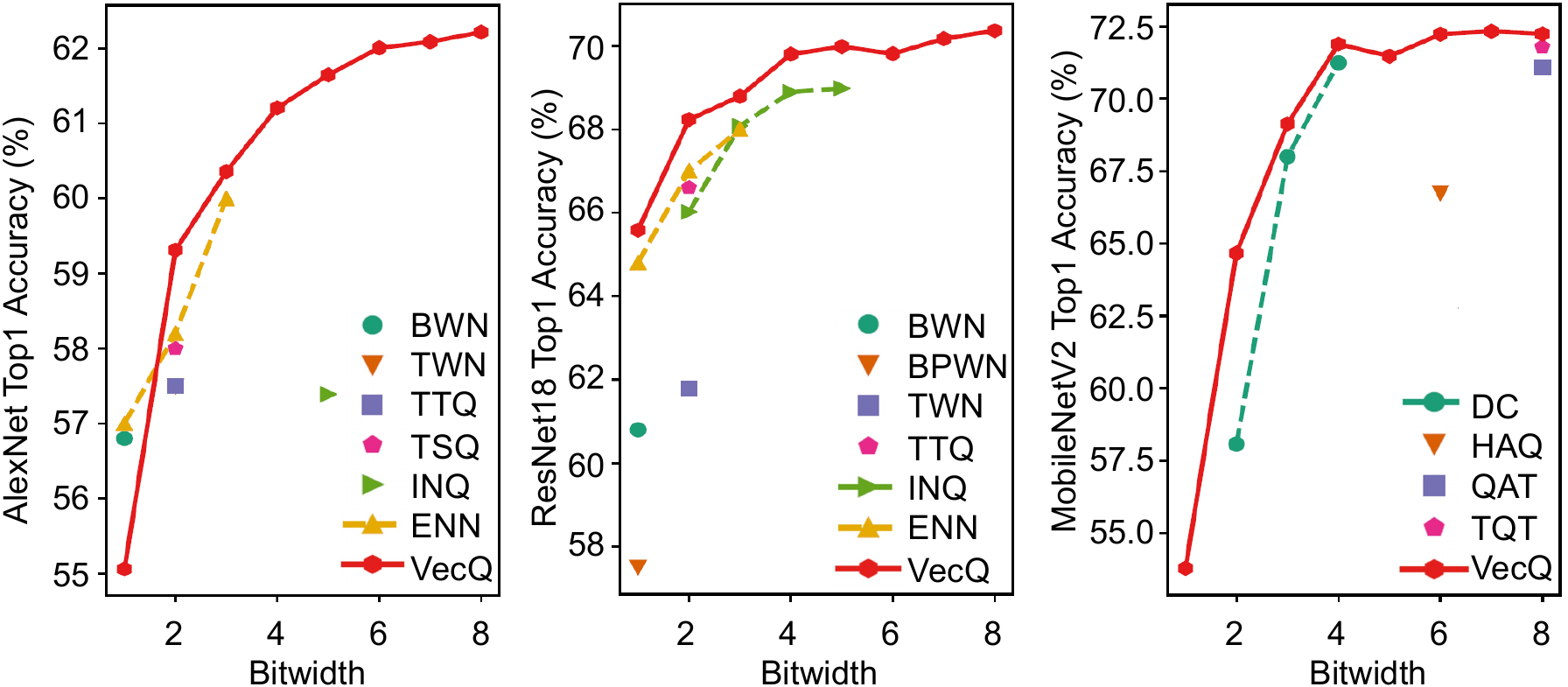}
    \caption{Comparison with state-of-the-art solutions.}
    \label{fig:vecq2sota}
\end{figure}

We evaluate VecQ on image classification task with the popular models and compare the results to the state-of-the-art quantization solutions with the same DNN model and bitwidth configurations. 
The state-of-the-art quantization solutions include BWN~\cite{rastegari2016xnor}, TWN~\cite{li2016ternary}, TTQ~\cite{zhu2016trained}, TSQ~\cite{TSQ2018}, INQ~\cite{INQ2017} and ENN~\cite{leng2018extremely}. Note here, not all of these quantization solutions provide bitwidth support from 1 to 8 bits.
As shown in Fig.~\ref{fig:vecq2sota}, our VecQ quantization outperforms most of the solutions with the same bitwidth configurations, and VecQ provides a wider range of bitwidth coverage as well. It only loses the advantage when comparing to the solutions specifically designed for binary weights.



\section{Efficient accelerator design and workload mapping}
\label{sec:compilers}

As discussed before, there exists an ever-widening barrier between fast DNN model design in software and slow hardware accelerator implementation.
To bridge the hardware-software gap, in this section, we introduce DNNBuilder \cite{zhang2018dnnbuilder} and PyLog \cite{huang2021pylog} to provide efficient solutions for automatically generating high-performance hardware accelerators for DNN workload deployments.

\subsection{DNNBuilder}
DNNBuilder is an end-to-end automation framework that can transform DNN designs from popular deep learning frameworks to highly optimized hardware deployment on customized accelerators implemented on FPGAs. Users are no longer required to design and optimize accelerators manually but can enjoy the auto-generated hardware accelerators for desired AI workloads. DNNBuilder introduces two major architecture innovations: the fine-grained layer-based pipeline architecture and the column-based cache scheme, which achieve 7.7$\times$ and 43$\times$ reduction of latency and on-chip memory usage, respectively. 
This subsection presents the novel designs introduced by DNNBuilder and showcases its promising edge AI performance.

\subsubsection{An end-to-end automation flow}

\begin{figure}[h]
  \centering
  \includegraphics[width=1\columnwidth]{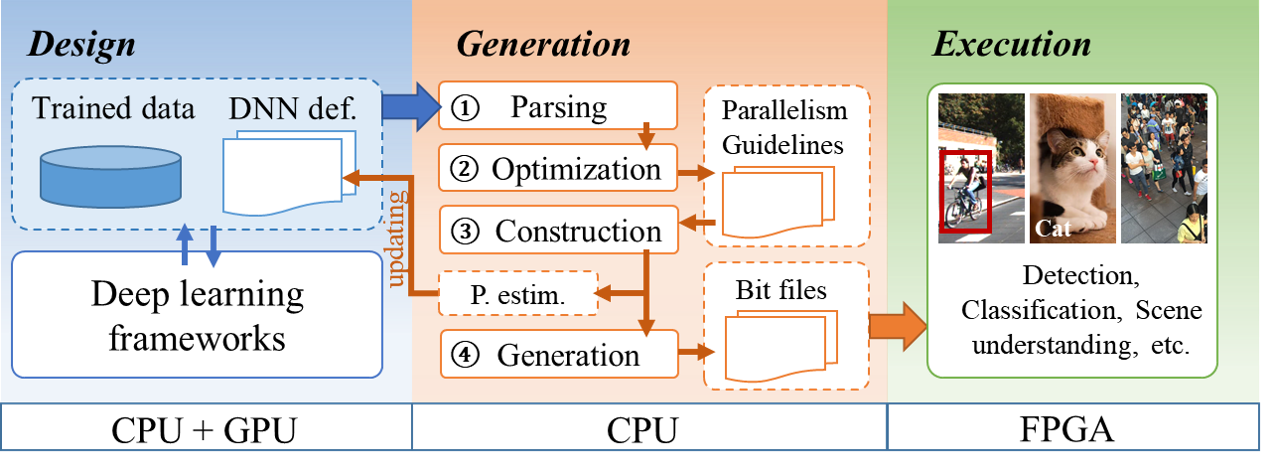}
  \caption{The end-to-end design flow introduced by DNNBuilder \cite{zhang2018dnnbuilder}.}
  \label{fig:dnnbuilder_design_flow}
\end{figure}

DNNBuilder produces customized DNN accelerators in three steps as \textit{Design}, \textit{Generation}, and \textit{Execution} (Fig. \ref{fig:dnnbuilder_design_flow}). 
During the \textit{Design} step, a DNN is designed and trained using deep learning frameworks, which in general employ CPUs and GPUs. After training, network definition files and trained parameters are passed to the next step. To ensure design freedom specified by users, the proposed flow supports hybrid quantization schemes, where different quantization schemes can be applied to the parameters and activations of different network layers, to explore tradeoffs among inference accuracy, resource utilization, performance, etc. One important feature of this step is the feedback function that provides hardware metrics estimation. If the current DNN runs slower or consumes more resources than expected, users could update their network designs, such as adjusting quantization schemes or modifying network layers to meet performance and resource requirements. This function also makes the hardware-software co-design possible.

In the \textit{Generation} step, 
\textbf{network parsing} is launched to decompose the input models. 
Different network layers, e.g., CONV, Pooling, and FC layers, are decomposed and then mapped to our pre-built RTL IPs, which are the basic building blocks of the generated accelerator. The computational intensive nested loops are captured by parameterized compute engines. 
Then, \textbf{automated optimization} works for exploring the hardware design space and provides configuration guidelines so that the generated accelerator can achieve maximum performance. Following these guidelines, \textbf{network construction} is responsible for building DNN implementations with the pre-built RTL IPs, dataflow controller, and memory instances, which are highly configurable to ensure the adaptability and scalability for various DNNs. After that, \textbf{code generation} generates accelerator related files for FPGA-based instances. 

In the \textit{Execution} step, the DNN accelerator is instantiated in FPGA with unified interfaces, including a FIFO-like data input/output interface and a weight access interface connecting the off-chip memory controller. In this final step, the DNN accelerator is ready for eventual deployment.

\subsubsection{Architecture novelties}

\begin{figure}[h]
  \centering
  \includegraphics[width=1\columnwidth]{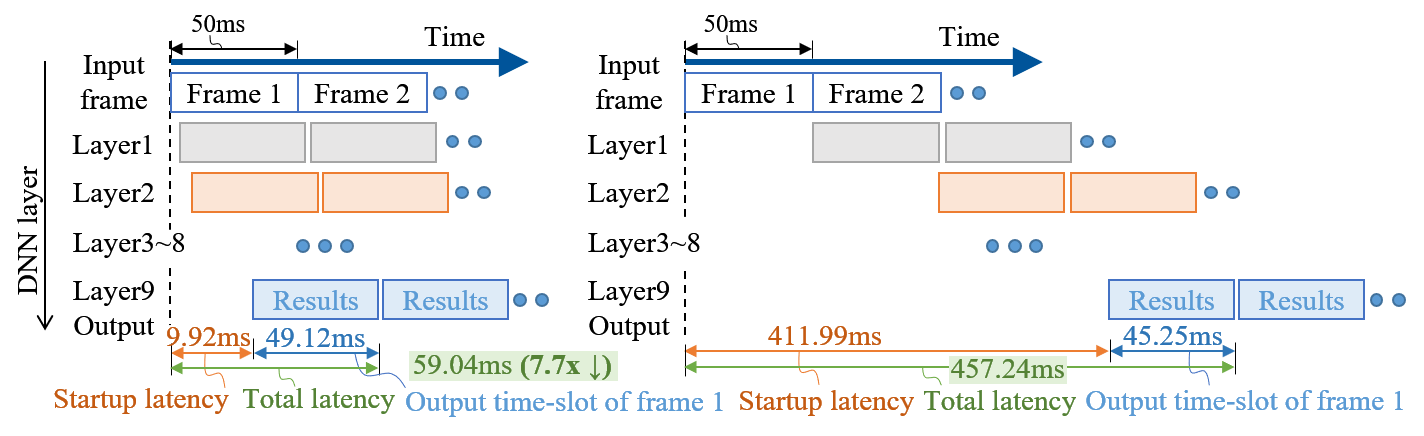}
  \caption{Latency comparison between the proposed fine-grained (left) and conventional (right) pipeline when handling the same object detection DNN model with a ZC706 embedded FPGA \cite{zhang2018dnnbuilder}.}
  \label{fig:dnnbuilder_latency}
\end{figure}

We propose a fine-grained layer-based pipeline to deliver high throughput performance and promising real-time response. 
Each major neural network layer, such as CONV or FC layer, in the targeted DNN model, is handled by one pipeline stage, as major layers dominate computation and memory consumption. The rest of the layers, such as batch normalization (BN), scale, and activation layers, are aggregated to their neighboring major layers so that we reduce the number of pipeline stages for lower latency. 
In addition, DNNBuilder enables pipeline stage overlapping to overcome the long initial latency, which is frequently encountered by conventional pipelines.
We demonstrate the proposed fine-grained layer-based pipeline by accelerating an object detection DNN model called YOLO \cite{redmon2017yolo9000} and show the results in Fig. \ref{fig:dnnbuilder_latency}.
DNNBuilder can effectively hide the data transmission delay and generate outputs even when the first input frame is still loading. It helps achieve a 7.7$\times$ smaller startup latency (9.92ms) compared to the conventional pipeline design (411.99ms).

\begin{figure}[h]
  \centering
  \includegraphics[width=0.72\columnwidth]{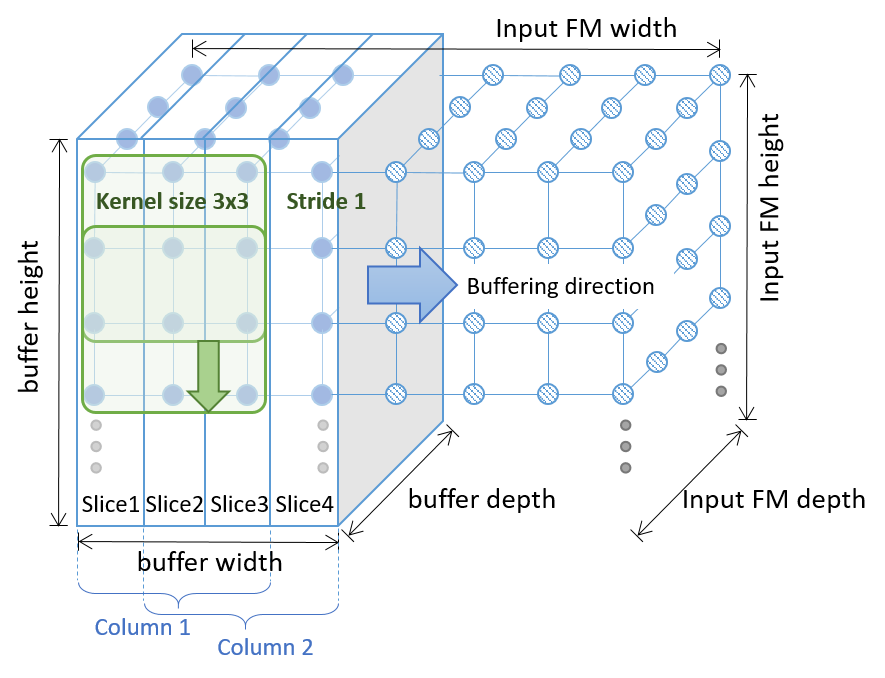}
  \caption{The proposed column-based cache scheme \cite{zhang2018dnnbuilder}.}
  \label{fig:dnnbuilder_cache}
\end{figure}

The other novel design is the column-based cache scheme, which reduces on-chip memory utilization during DNN inference and supports high-definition image input for resource-constrained embedded systems. 
By following the pipeline architecture, intermediate results between pipeline stages are stored on-chip to guarantee seamless pipeline operations. However, feature maps can be enormous when inputs become large in real life and become impossible to be held on-chip entirely.
The column-based cache scheme is designed to address this problem as it only keeps a subset of the input feature map on chip. 
Fig. \ref{fig:dnnbuilder_cache} shows an example when DNNBuilder processes a convolution layer (with kernel size$=$3 and stride$=$1). Since slices 1$\sim$3 contribute to the first sliding window operation (from top to bottom), we name the first three slices as column 1. Similarly, column 2 represents the amount of data for the second sliding window operation, so that slices 2$\sim$4 constitute column 2. DNNBuilder caches at least two columns before starting computing, which allows the kernel to perform the second vertical sliding window operation immediately after finishing the first one. Delay caused by data shortage will not happen by caching one more column. Meanwhile, slice 5 will start buffering to form the next column (with slices 3$\sim$5) after releasing the room taken by slice 1. 
%
By serving the same objection detection AI model (YOLO with high-definition inputs), the proposed column-based cache can significantly reduce 43$\times$ on-chip memory usage compared to the accelerator without this technology \cite{zhang2018dnnbuilder}.

\subsubsection{State-of-the-art performance}

We demonstrate our design by accelerating popular AI workloads on an embedded platform (ZC706). As shown in Table \ref{tab:dnnbuilder_comp_embedded}, our DNNBuilder generated design reaches the best performance (524 and 262 GOPS in Fix8 and Fix16 quantization schemes) and power efficiency (72.8 GOPS/Watt in Fix8 and 36.4 GOPS/Watt in Fix16).
We also extend our comparison to the embedded GPU (TX2) in Table \ref{tb:dnnbuilder_comp_gpu_cpu}. The DNNBuilder-generated design can deliver higher efficiency than the TX2-based solution even without using batch processing (batch size = 1), and it can achieve up to 47.2 image/Second/Watt.

\begin{table}[h]
\footnotesize
\caption{Comparison with existing embedded FPGA-based DNN accelerators \cite{zhang2018dnnbuilder}.}
\label{tab:dnnbuilder_comp_embedded}
\begin{center}
\newcommand{\tabincell}[2]{\begin{tabular}{@{}#1@{}}#2\end{tabular}}
\begin{tabular}{|c|c|c|c|}
\hline
Reference & \cite{qiu2016going}& \cite{xiao2017exploring} & DNNBuilder \\ 
\hline
FPGA chip & Zynq XC7Z045 & Zynq XC7Z045 & Zynq XC7Z045 \\
\hline
Frequency & 150 MHz & 100 MHz &200MHz \\
\hline
Network & VGG&VGG& VGG\\
\hline
Precision & Fix16 & Fix16 & Fix16 (Fix8)\\
\hline
\tabincell{c}{DSPs (used/total)} & 780/900 & 824/900 & 680/900\\
\hline
DSP Efficiency & 44.0\%& 69.6\% & 96.2\% \\
\hline
\tabincell{c}{Performance (GOPS)} & 137 & 230 & 262 (524)\\
\hline
\tabincell{c}{Power Efficiency (GOPS/W)} & 14.2& 24.4& 36.4 (72.8)\\
\hline
\end{tabular}
\end{center}
\end{table}

\begin{table}[h]
\footnotesize
\caption{Alexnet inference comparison on embedded GPU and FPGA platforms \cite{zhang2018dnnbuilder}.}
\label{tb:dnnbuilder_comp_gpu_cpu}
\begin{center}
\newcommand{\tabincell}[2]{\begin{tabular}{@{}#1@{}}#2\end{tabular}}
\begin{tabular}{|c|c|c|c|c|c|}
\hline
Platform & Precision & Batch & \tabincell{c}{Throughput\\(img./S)} & \tabincell{c}{Power\\(W)} & \tabincell{c}{Efficiency\\(img./S/W)} \\ \hline
DNNBuilder (ZC706) & Fix16, Fix8 &1, 2& 170, 340	& 7.2 & 23.6, 47.2\\ \hline
GPU-TX2\cite{franklin2017nvidia} & Float16 &2& 250	& 10.7 & 23.3\\ \hline 
\end{tabular}
\end{center}
\end{table}

\subsection{PyLog: A Python-based FPGA Programming Flow}

The fast-growing complexity of new applications and new use scenarios poses serious challenges for computing systems. Embedded hardware accelerator systems have demonstrated great flexibility, performance, and efficiency in many different applications and scenarios. However, as system complexity and application complexity grow rapidly, programming and optimizing embedded accelerator systems require great manual efforts and consume a lot of time. Compiling and optimizing a general application specified in high-level programs like Python are becoming common tasks in creating embedded accelerator designs. High-level Synthesis (HLS) transforms design inputs written in high-level languages (e.g., C++, OpenCL, Python) to hardware descriptions in RTL (Register-Transfer Level) languages such as Verilog. HLS offers up to $10\times$ code reduction and $1,000\times$ simulation time reduction over manual RTL design solutions. HLS has been intensively studied in the past three decades~\cite{papakonstantinou2009fcuda, rupnow2011study,  papakonstantinou2011multilevel, fcudanoc1, fcudanoc, fcudasoc, fcudahb, jasoncontrolflow, jasonhlsfpga, autopilot, zhiruplatform, legup2011, legup2013, scalehls, chen2009lopass, lowpowerAssignment1, lowpowerAssignment2}, and there are popular commercial HLS tools used by many designers~\cite{xilinxvitis, mentorhls}.

\subsubsection{PyLog Flow Overview}

PyLog \cite{huang2021pylog} is a Python-based high-level programming flow for FPGAs. It allows users to create FPGA accelerators with Python code using PyLog high-level operators and Python syntax. PyLog presents a unified programming model for host and accelerator logic with consistent Python-based syntax and semantics. This seamless host-accelerator programming model enables agile system design, convenient functional simulation, and flexible design space exploration.  

\begin{figure}[ht]
    \centering
    \includegraphics[width=0.7\textwidth]{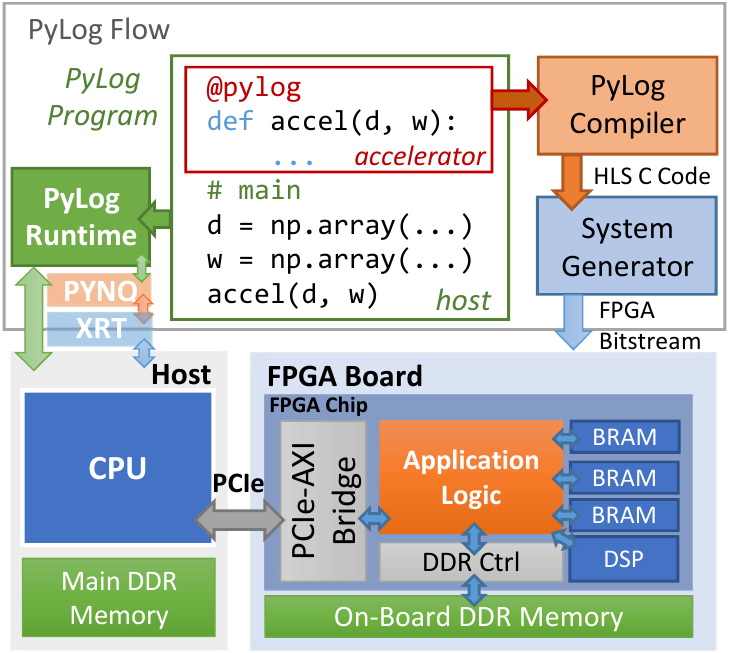}
    \caption{The PyLog Flow and Example System Architecture \cite{huang2021pylog}}
    \label{fig:pylog_overview}
\end{figure}

Fig. \ref{fig:pylog_overview} shows the overall PyLog at high level. PyLog flow allows users to create efficient FPGA accelerators and program host system with Python. The input to the PyLog flow is Python code, where the FPGA kernel function is decorated with the \texttt{@pylog} decorator. The PyLog flow contains an accelerator synthesis flow and a runtime flow.

In the accelerator synthesis flow, the \texttt{@pylog} decorator calls the PyLog compiler to compile the kernel function into optimized high-level synthesis (HLS) C code, which is then compiled into efficient FPGA IPs with HLS flow, and integrated into a complete accelerator system by PyLog system generator. Beside the PyLog kernel function, the rest of the PyLog program is interpreted by the standard Python interpreter running on the host CPU side, which supports all Python libraries and language features. This part of PyLog program naturally becomes the host program of whole accelerator system. After the accelerator system is generated by the synthesis flow, the system can be deployed at the target FPGA platform using the generated FPGA bitstream configuration file, and runs with support from the PyLog runtime flow. During runtime, PyLog runtime can prepare input data, launch accelerator, and collect results according to the host code. Host CPU and the FPGA accelerator interactions are handled automatically by the PyLog runtime and the underlying Xilinx PYNQ library \cite{pynq}.

\subsubsection{PyLog Features}

PyLog has several unique features that help users to create FPGA accelerators more efficiently. 

\begin{enumerate}[(i)]

\item 
\textbf{High-Level Operators.} 

In addition to commonly used Python standard language features, PyLog also supports several built in high-level operators and NumPy operators that allow users to express computation patterns at high level and enable better compiler optimizations. Table \ref{tb:pylog_features} summarizes the language features supported in PyLog, including PyLog high-level operators, NumPy operators, and standard Python features. Listing \ref{code:map} demonstrates a few example usages of PyLog \texttt{map} and \texttt{dot} operators.

\begin{table}[t]
\caption{{PyLog Supported Language Features \cite{huang2021pylog}}}
\begin{center}
\begin{tabular}{|l|l|} \hline
\textbf{Category} & \textbf{Operators} \\ \hline
PyLog high-level operators & \texttt{map}, \texttt{dot}, user-defined ops  \\ \hline
\multirow{2}{*}{NumPy operators}  & \texttt{argmax}, \texttt{argmin}, \texttt{max}, \texttt{min}, \texttt{matmul}, \\ & \texttt{convolve}, \texttt{sort} \\ \hline
\multirow{3}{*}{Python features} 
& \texttt{list}, \texttt{functions}, \texttt{calls}, \texttt{lambda},  \\
& \texttt{for}, \texttt{while}, \texttt{if...else...},  \texttt{slice},\\ 
& \texttt{subscript}, \texttt{attribute}, \texttt{bin\_op},  \\
& \texttt{unary\_op}, \texttt{return} \\
\hline
\end{tabular}
\label{tb:pylog_features}
\end{center}\vspace{-10pt}
\end{table}

\begin{pythonenv}[caption={PyLog \texttt{map} and \texttt{dot} examples. \cite{huang2021pylog}}, label={code:map}, emph={map, dot}]
# Vector add
out = map(lambda x, y: x + y, vec_a, vec_b)

# 1D convolution
out = map(lambda x:w0*x[-1]+w1*x[0]+w2*x[1], vec)

# Inner product
out_vec[i] = dot(matrix[i,:], in_vec)

# Square matrix multiplication
out = map(lambda x,y: dot(x[0,:],y[:,0]), ma, mb)
\end{pythonenv}

These operators not only simplify programming for users, they also pass more information on computation to the compiler (e.g., computation patterns, data flow information, data/task parallelism, etc.), compared to programming in C/C++, and thus allows compilers to perform more optimizations and choose the optimal code generation.

\begin{figure}[!t]
\centering
\includegraphics[width=0.8\linewidth]{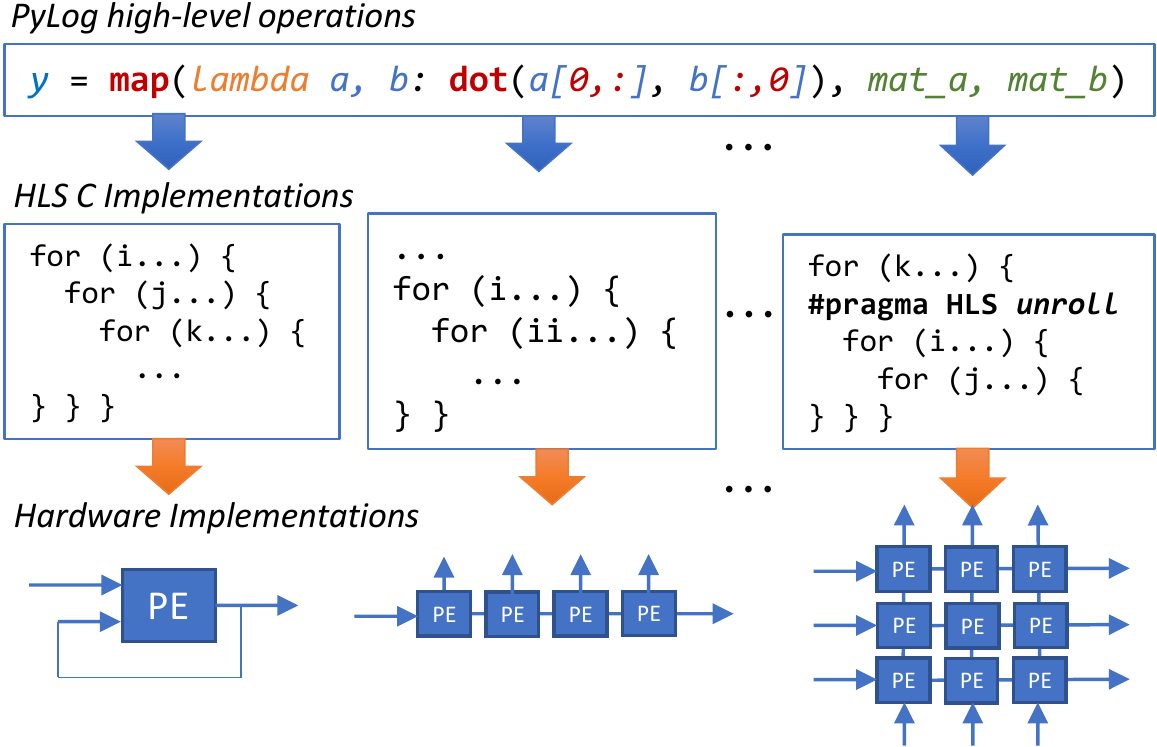}
\caption{Different implementations generated from the same PyLog code. \cite{huang2021pylog}}
\label{fig:gen_arches}
\end{figure}

Fig. \ref{fig:gen_arches} shows an example of generating multiple hardware implementations from a PyLog \texttt{map} operation. The compiler generates HLS C implementations in different styles, which corresponds to different hardware structures, e.g. shift registers, systolic arrays, etc. Depending on the context and constraints, the optimal implementation will be chosen. 

\item
\textbf{Type Inference and Type Checking.}

Python is a dynamically typed languages and there is no explicit type declaration in the Python code. PyLog has a builtin type inference and type checking engine that can infer the type and shape information of code objects in the PyLog kernel functions. This type inference engine is critical in PyLog since same operators may have completely different meanings when applied to operands with different types or shapes. With this type inference engine, PyLog users do not need to provide explicit type annotations or hints in PyLog program.

\item
\textbf{Compiler Optimizations.}

PyLog provides a set of compiler optimizations that improve the design quality of generated accelerators. PyLog uses its own PyLog intermediate representation (PLIR) as the internal representation of the input code. PyLog code analysis and transformation passes work on PLIR to perform a sequence of optimizations including high-level operator lowering, loop transformation, HLS pragma insertion, etc. The internal PLIR is capable of expressing different design options and can therefore form a design space that not only covers low-level design parameter tuning, but also high-level design pattern selection, which has not been explored in previous tools. 

\end{enumerate}

\subsubsection{PyLog Evaluation Results}

We evaluate the performance of PyLog in terms of expressiveness and accelerator performance, using real-world applications. 

\begin{enumerate}[(i)]

\item 
\textbf{Expressiveness. }We evaluated the expresiveness of PyLog by comparing the number of lines of code to implement a benchmark using PyLog and HLS C. 

\begin{figure}[!t]
\centering
\includegraphics[width=0.8\linewidth]{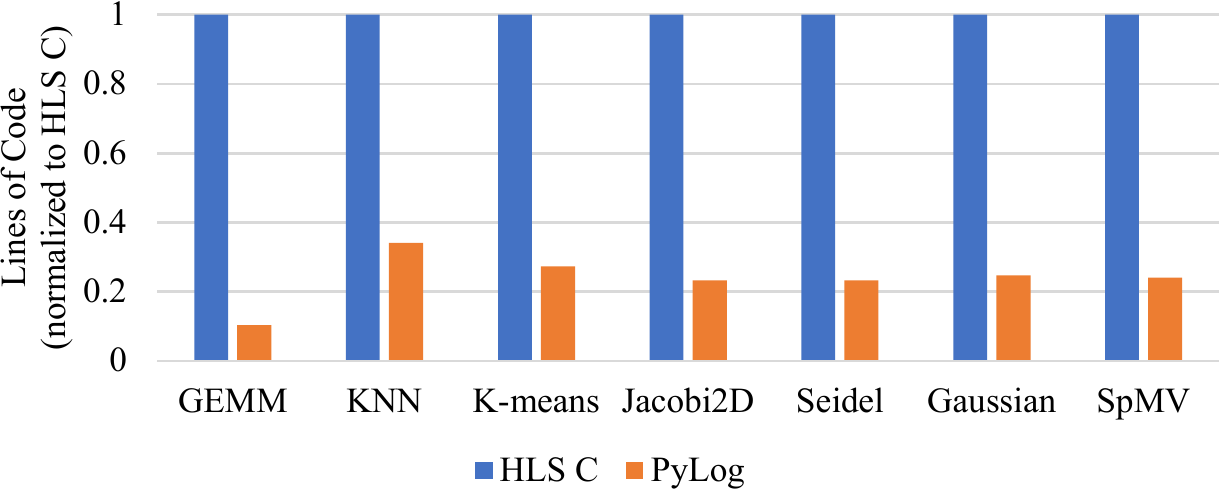}
\caption{Length of HLS C code and PyLog code. \cite{huang2021pylog}}
\label{fig:code_length}
\end{figure}

For the benchmarks evaluated, on average PyLog only needs around $30\%$ of the code length of HLS C. This indicates that PyLog provides good expressiveness compared with HLS C and allows users to describe their computation with fewer lines of code. 

\item 
\textbf{Accelerator Performance. }
We evaluated the performance of the accelerators generated by PyLog using real-world benchmarks. Evaluation was done on Amazon EC2 F1 f1.2xlarge instance. The evaluated benchmarks are from different domains and have various computation patterns. They are representative FPGA workloads, including linear algebra, data analytics, stencil, sparse operations, etc. Amazon EC2 F1 f1.2xlarge instance is a cloud computing platform that has an 8-core Intel Xeon E5-2686 v4 CPU and a Xilinx Virtex UltraScale+ XCVU9P FPGA. Table \ref{tb:pylog_results_aws} shows the evaluation results. The table lists the FPGA resource utilization as well as the accelerator execution time. We compared the PyLog accelerator execution time against the optimized CPU time as well as the execution time of accelerators generated from \cite{heterocl}. On average, PyLog accelerators achieve around $3.17\times$ and $1.24\times$ speedup over CPU baseline and manually optimized accelerators \cite{huang2021pylog}.

\begin{table*}[!t]
\small
\caption{Accelerator Performance Evaluation on AWS F1 Instance}
\begin{threeparttable}
\begin{center}
\begin{tabular}{|r|r|r|r|r|c|c|ccc|cc|} \hline
\textbf{Benchmark} & \textbf{LUT}  & \textbf{FF}  & \textbf{BRAM}  & \textbf{DSP}  & {$f$(MHz)}  & {$P$(W)}   & {$T_\text{CPU}$} & $T_\text{HCL}$  & $T_\text{PyLog}$ &  $\frac{T_\text{CPU}}{T_\text{PyLog}}$  & $\frac{T_\text{HCL}}{T_\text{PyLog}}$\\
\hline
KNN & 109276 & 74889 & 425 & 0 & 256.40 & 37.222 & 0.48 & 0.45 & 0.26 & 1.85 & 1.73 \\ 
K-means & 10829 & 17604 & 3 & 7 & 273.97 & 37.429 & 38.16 & 4.24 & 4.45 & 8.58 & 0.95 \\ 
Jacobi~\cite{grauer2012tuning} & 93925 & 111144 & 96 & 704 & 269.03 & 37.327 & 11.31 & 8.25 & 5.19 & 2.18 & 1.59 \\ 
Seidel~\cite{grauer2012tuning} & 47304 & 57854 & 30 & 304 & 269.03 & 37.341 & 21.37 & 8.22 & 5.16 & 4.14 & 1.59 \\ 
Gaussian~\cite{grauer2012tuning} & 56580 & 75846 & 48 & 688 & 147.15 & 37.783 & 23.63 & 7.34 & 5.19 & 4.55 & 1.41 \\ 
GEMM & 12868 & 63759 & 655 & 1024 & 250.00 & 39.657 & 60.34 & 8.13 & 13.05 & 4.62 & 0.62 \\
SpMV & 8294 & 12787 & 25 & 21 & 273.97 & 37.225 & 0.29 & - & 0.24 & 1.21 & - \\ 
Histogram~\cite{kastner2018parallel} & 4096 & 7647 & 13 & 0 & 273.97 & 37.327 & 5.85 & - & 2.07 & 2.83 & - \\ \hline
\multicolumn{10}{|r|}{Geometric Mean} &  \textbf{3.17} & \textbf{1.24} \\
\hline
\end{tabular}
\begin{tablenotes}\footnotesize
\item $T_\text{CPU}$: Execution time on CPU; $T_\text{HCL}$: Execution time on HeteroCL\cite{heterocl} generated accelerator; $T_\text{PyLog}$: Execution time on PyLog generated accelerator; All time values are in milliseconds (ms); `-' means the implementation is not publicly available. 
\end{tablenotes}
\end{center}
\end{threeparttable}
\label{tb:pylog_results_aws}
\end{table*}

\end{enumerate}

\section{Efficient Optimizations}
\label{sec:optimization}

With a great number of efficient optimization techniques, in this section, we introduce three key optimization techniques: hardware-aware NAS, FPGA/DNN co-design, a specialized approach for FPGA/DNN co-design~\cite{hao2019fpga}, and a unified differentiable co-design approach, across different platforms~\cite{li2020edd}.

\subsection{Overview of Hardware-aware Neural Architecture Search (NAS)}
\label{sec:hardware_aware_NAS}

\begin{table}[]

\caption{A brief overview of Neural Architecture Search components and example algorithms.}\label{tab:NAS cat}
\begin{center}

\begin{tabular}{|ll|l|}

\hline
\multicolumn{1}{|l|}{\multirow{4}{*}{Search space}}     & \multirow{2}{*}{Architecture search space} & NASNet~\cite{zoph2018learning}, DARTS~\cite{liu2018darts}, FB-Net~\cite{wu2019fbnet},                           \\
\multicolumn{1}{|l|}{}                                  &                                            & ProxylessNAS~\cite{cai2018proxylessnas}, FlexiBERT~\cite{tuli2022flexibert}...                       \\ \cline{2-3} 
\multicolumn{1}{|l|}{}                                  & \multirow{2}{*}{Hardware search space}     & Quantization, sparsification, tiling parameters, \\
\multicolumn{1}{|l|}{}                                  &                                            & number of PEs, other HW specific parameters...    \\ \hline\hline
\multicolumn{1}{|l|}{\multirow{4}{*}{Search algorithm}} & \multirow{2}{*}{Sampling-based}              & Reinforcement learning, evolutionary algorithm,  \\
\multicolumn{1}{|l|}{}                                  &                                            & Bayesian optimization, random search...          \\ \cline{2-3} 
\multicolumn{1}{|l|}{}                                  & \multirow{2}{*}{Supernet-based}                  & DARTS, Random sampling~\cite{li2020random}, SNAS~\cite{xie2018snas},               \\
\multicolumn{1}{|l|}{}                                  &                                            &  EDD~\cite{li2020edd}, ProxylessNAS, OFA~\cite{cai2019once}...                             \\ \hline \hline
\multicolumn{2}{|l|}{\multirow{2}{*}{Network evaluation}}                                            & Early stopping, NAO~\cite{luo2018neural}, NASWOT~\cite{mellor2021neural},         \\
\multicolumn{2}{|l|}{}                                                                               & Synflow~\cite{abdelfattah2021zero},    GenNAS~\cite{li2021generic}...                                        \\ \hline
\end{tabular}
\end{center}
\end{table}

Neural Architecture Search (NAS) refers to the automated process of neural architectural design~\cite{kyriakides2020introduction}. It has been largely successful in producing many state-of-the-art networks.
Typically, a NAS process requires three distinct components as shown in Table~\ref{tab:NAS cat}:
\begin{enumerate}
\item \textbf{Search space}. A search space includes all possible network architectures that follow a predefined template. For example, the networks can be sequential layer-wise architecture~\cite{zoph2016neural,baker2016designing}, cell-based architecture~\cite{zoph2018learning}, and  hierarchical architecture~\cite{tan2019mnasnet}. Also, hardware parameters should be considered into the search space for HW-awared NAS.

\item \textbf{Search algorithm}. The fundamental component of NAS is the search algorithm. Given the prohibitively large search space, the search algorithm can greatly influence the efficiency of the search and the effectiveness of the final network architecture. Generally, search algorithms can be classified into two categories: supernet-based search~\cite{liu2018darts, wu2019fbnet} and sampling-based search~\cite{zoph2016neural, zoph2016neural, luo2018neural, li2021generic}. 

\item \textbf{Network evaluation}. Network evaluation is the key for efficient NAS, since fast evaluation is required to estimate the quality of individual networks to guide the search algorithm to choose top-performing architectures from the search space. Network evaluation can be prohibitively expensive due to network training, so that various approximation approaches have been proposed to expedite the evaluation such as few-shot and one-shot training~\cite{zhao2021few, li2019finding} and using proxy tasks~\cite{abdelfattah2021zero}. 
\end{enumerate}

\subsection{HW-aware NAS Formulation}
In recent years, driven by the need of deploying power-hungry DNNs into resource-constrained devices, hardware-aware NAS (HW-NAS) has emerged as one of the most promising techniques~\cite{benmeziane2021comprehensive}.
There is a great amount of hardware-aware work, each of which often adopts a specific hardware device (CPU, GPU, embedded/mobile device) and requires a different hardware-cost metric (e.g., prioritizes latency or energy). For example, FBNet~\cite{wu2019fbnet} develops a
differentiable neural architecture search (DNAS) framework and discovers state-of-the-art DNNs balancing both accuracy and hardware efficiency, by incorporating a loss consisting of both the cross-entropy loss that leads to better accuracy and the latency loss that penalizes the network’s
latency on a target device. To provide more integrated co-optimization solutions, EDD~\cite{li2020edd} fuses the design space of DNN architecture and hardware accelerator and formulates the DNN and hardware design as a co-search problem. EDD aims to discover the most suitable combination of DNN and hardware within the co-search space and maximize software and hardware metrics given the targeted edge AI application. Once for All (OFA)~\cite{cai2019once} is the first work that proposes an elastic training scheme for supernet. By training the supernet, high-accuracy architectures is directly searched by selecting from the OFA network without additional training.

One of the classic search method for HW-NAS is to first define a template based search space, and then incorporate hardware performance into the loss function:
\begin{equation}
    \mathcal{L} = \mathcal{L}_{T} + \mathcal{L}_{HW} ~~~~~~~ \text{or}~~~~~~ \mathcal{L} = \mathcal{L}_{T} \cdot \mathcal{L}_{HW}
    \label{eq:hw-nas-obj}
\end{equation}
where $\mathcal{L}_{T}$ is the task-specific loss 
of NAS, such as cross-entropy loss for classification tasks or Mean squared error (MSE) loss for regression tasks. $\mathcal{L}_{HW}$ is the hardware performance loss, such as measured or estimated execution latency of the network architectures on the target device.


\subsection{FPGA/DNN Co-design} \label{sec:fpga-dnn-codesign}

\begin{figure}
    \centering
    \includegraphics[width=0.9\textwidth]{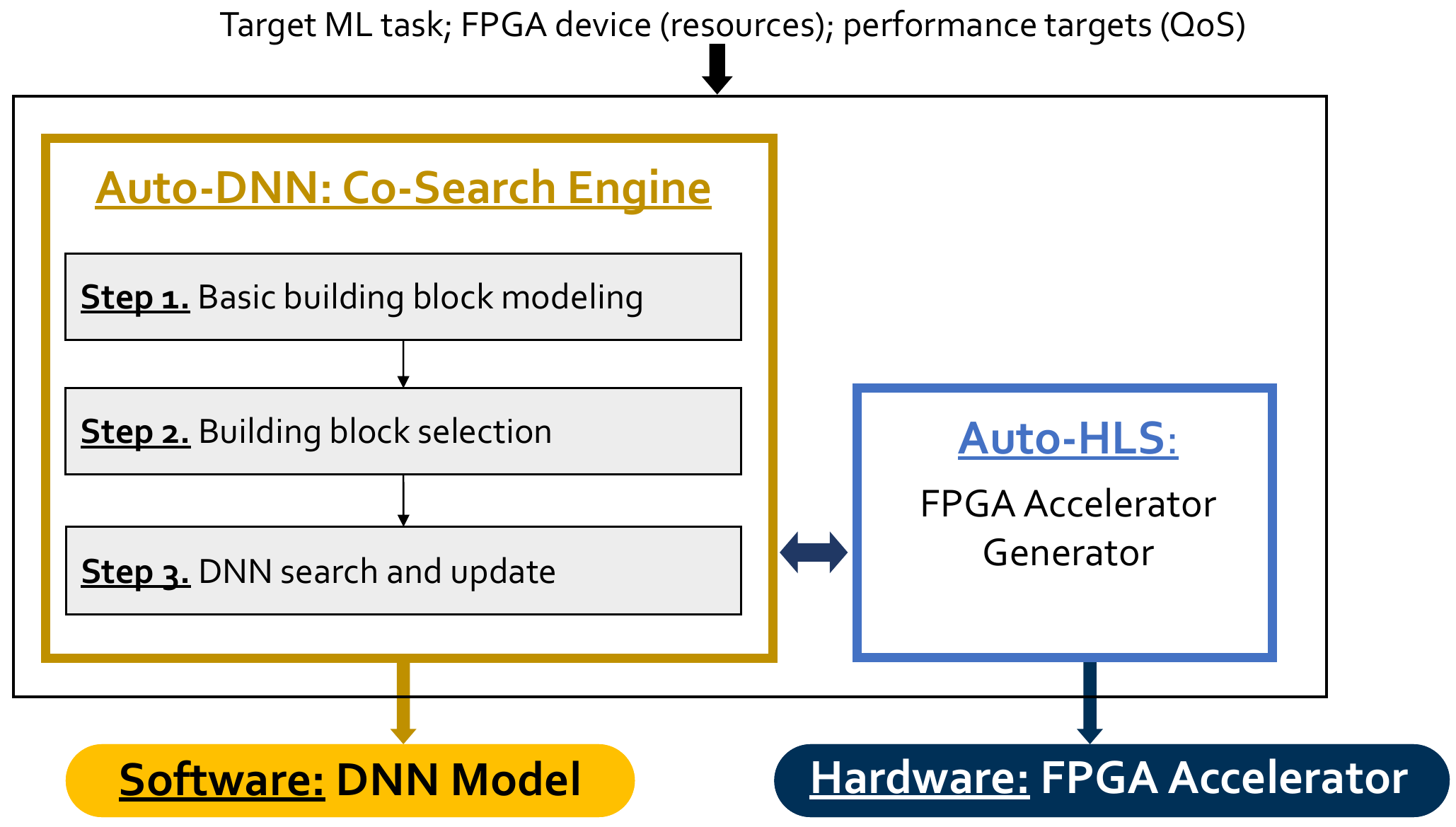}
    \caption{FPGA/DNN co-design framework~\cite{hao2019fpga}.}
    \label{fig:auto-dnn-auto-hls}
\end{figure}

Hao and Chen first proposed the concept of accelerator and DNN co-design in an invited paper titled ``Deep Neural Network Model and FPGA Accelerator Co-design: Opportunities and Challenges''~\cite{hao2018deep}, where they advocated ``automatically generate both DNN models and their corresponding implementations as pairs''. 
Later, based on the proposed co-design method, we implemented the first simultaneous FPGA/DNN co-design framework~\cite{hao2019fpga}. It has two major components, as shown in Fig.~\ref{fig:auto-dnn-auto-hls}: (1) a hardware-oriented bottom-up DNN model design, named \textit{Auto-DNN}, which is an efficient search engine to explore DNN candidates under hardware resource and performance constraints; (2) a DNN-driven top-down FPGA accelerator design, named \textit{Auto-HLS}, which is a fast board-level design generator to automatically map DNNs onto FPGAs.

\subsubsection{The key to co-design: Bundle}

\begin{figure}
    \centering
    \includegraphics[width=0.85\textwidth]{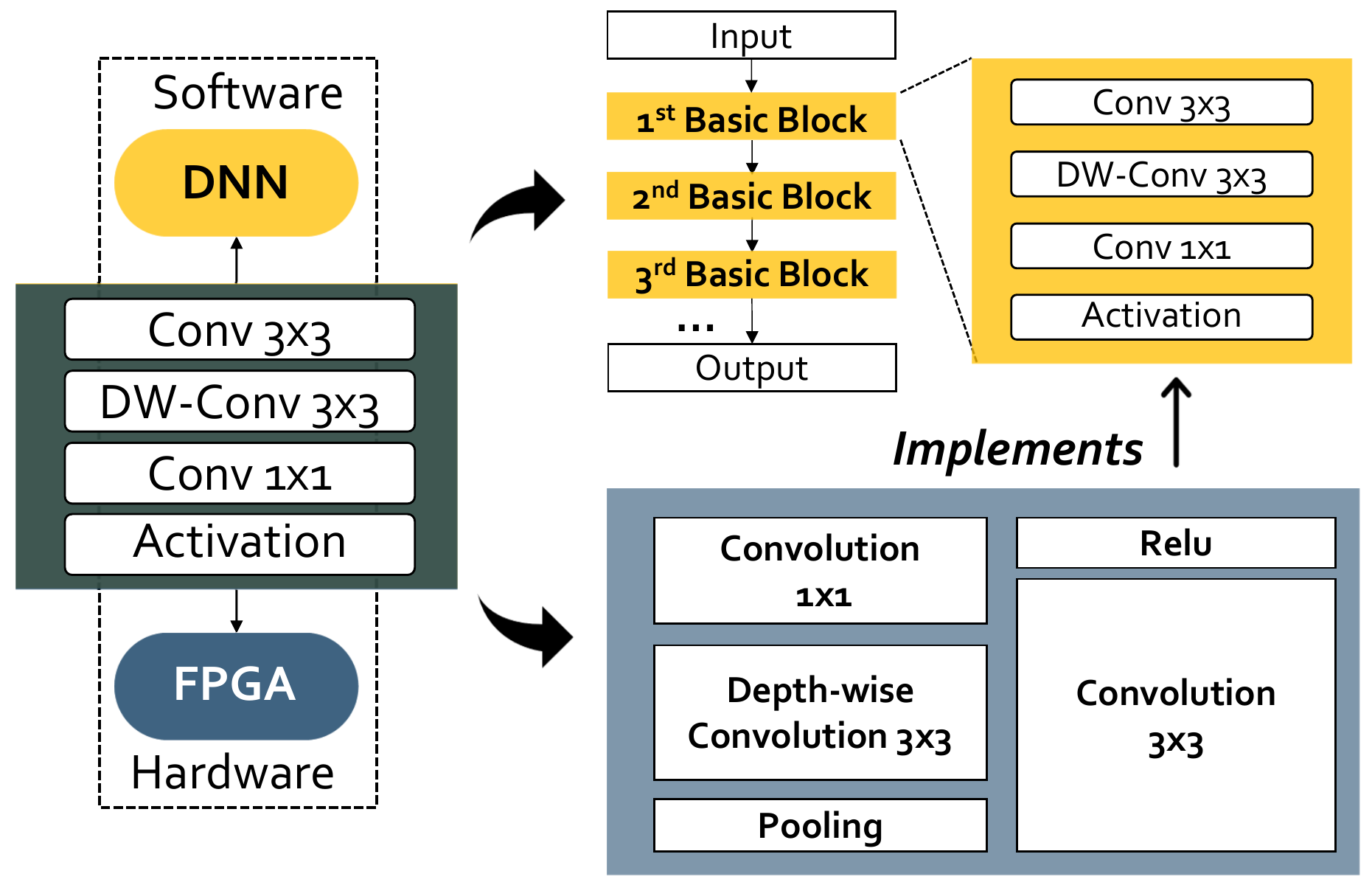}
    \caption{The key to co-design: Bundle -- the common basic building block to both DNN design and accelerator design~\cite{hao2019fpga}.}
    \label{fig:bundle}
\end{figure}

The key to achieve co-design, i.e., to execute \textit{Auto-DNN} and \textit{Auto-HLS} simultaneously, is to propose basic building blocks that can be used to construct both DNNs and their accelerators at the same time. We call such building blocks \texttt{Bundles}, the common building block of both DNN architectures as well as their hardware implementation, as shown in Fig.~\ref{fig:bundle}. The benefits of \texttt{Bundles} are two-fold.
First, a DNN can be constructed by replicating a bundle for a certain number of layers with pooling layers inserted, which is a common and effective way to construct high-quality DNNs, such as the residual block in ResNet~\cite{he2016deep}, the Inception block in GoogLeNet~\cite{szegedy2015going}; meanwhile, many NAS approaches follow such cell-based strategy~\cite{zoph2018learning, tan2019mnasnet, liu2018darts}.
Second, an accelerator can be constructed by building a hardware module for the certain bundle and reusing it for different DNN layers, given that the DNN is built by replicating the bundle; this can significantly reduce the resource usage of the accelerator by resource sharing and shorten the hardware development cycle.
As an example, a \texttt{Bundle} can be a set of DNN layers including: one $3\times 3$ convolution, one batch normalization, one activation, one $1\times 1$ convolution, and one activation.
Meanwhile, the hardware accelerator will need one instance for the $3\times 3$ convolution, one instance for the $1\times 1$ convolution, and so on.


\subsubsection{Progressively reducing search space}
It is non-trivial to select an optimal \texttt{Bundle} given the large design space and the prohibitively long DNN training time. Therefore, it it essential to narrow down the search space as early as possible. Our approach is in a three-step progressive way, by filtering out unfavourable \texttt{bundles} at early stage and conducting detailed search at later stage using promising ones.
The three steps are as follows.

\noindent
\textbf{Step 1}. Analytical models for performance and resource estimation for bundles and DNNs.
Denoting a Bundle as $bund_i$, the resource of $bund_i$ is computed as:
\begin{equation}
\label{eq:res-comp}
\vspace{-8pt}
    Res_{bund_i}^r = \sum_{p_j} Res^r_j + \Gamma_i^r
\vspace{-0pt}
\end{equation}
where $Res_j^r$ is the resource usage of instance $p_j$ of resource type $r$ ( including DSP, LUTs, FF and BRAM).
$\Gamma_i^r$ represents other resource overhead such as LUTs consumed by control logic and multiplexers.

The latency of a Bundle is estimated as:
\begin{equation} \label{eq:bund_lat}
\vspace{-8pt}
Lat_{bund_i} = \alpha_i \cdot \sum_{p_j} Comp_j + \frac{\beta_i \cdot \Theta(Data_i)}{bw}
\vspace{-0pt}
\end{equation}
where $Comp_j$ is the computation latency of instance $p_j$, and $\Theta(Data_i)$ is the data amount processed by $bund_i$. 
$bw$ represents the off-chip memory bandwidth.
Denote the latency of one execution of $p_j$ as $lat_j$, and the total number of reuses of $p_j$ as $reuse_j$, the computation latency $Comp_j$ is estimated as:
\begin{equation} \label{eq:bund-comp}
\vspace{-8pt}
    Comp_j = \sum_{1 \leq j \leq n} reuse_{j} \cdot lat_{j}
\vspace{+2pt}    
\end{equation}
$reuse_j$ can be computed by the input/output dimensions of the data processed by the IP and the data dimensions of $p_j$'s interface. The parameter $\alpha_i$ in Eq.~\ref{eq:bund_lat} describes how much computation is overlapped because of IP pipelining,
and $\beta_i$ describes how much data transfer is overlapped during computations.
$\alpha_i$, $\beta_i$ and $\Gamma_i$ will be determined for each $bund_i$ using \textit{Auto-HLS} sampling.

The overall DNN latency based on $Lat_{bund_i}$ in Eq. \ref{eq:bund_lat} is estimated as:
\begin{equation} \label{eq:DNN-overall-latency}
\vspace{-10pt}
    Lat_{DNN} = \sum_{i=1}^{N} Lat_{bund} + \phi \cdot Lat_{DM}
\vspace{+4pt}
\end{equation}
where $N$ is the the number of Bundle repetitions of the DNN, and $\phi \cdot Lat_{DM}$ represents the inter-bundle data movement latency.
For overall DNN resource utilization, we have:
\begin{equation} \label{eq:DNN-overall-resource}
\vspace{-8pt}
    Res_{DNN} = Res_{bund_i} + \gamma \cdot Res_{ctl}
\vspace{+2pt}
\end{equation}
where $Res_{bund_i}$ is the resource of $bund_i$, and $Res_{ctl}$ is additional control logic overhead, e.g., finite state machine and multiplexers.
$\phi$, $\gamma$, $Lat_{DM}$ and $Res_{ctl}$ will be decided and calibrated through actual hardware synthesis and implementation.

\noindent
\textbf{Step 2}. Bundle evaluation and selection.
In this step, we evaluate the latency, resource, and accuracy metrics for each Bundle, as defined in Step 1.
Since we cannot evaluate the accuracy for a single Bundle, we replicate a Bundle for $n$ times to build a DNN and train it for a small number of epochs (20 in the experiment).
We plot Pareto curves for the Bundles to examine the tradeoff between DNN accuracy and resource utilization, and the Bundles on the Pareto curve will be selected for detailed search in the next step.

\noindent
\textbf{Step 3}. DNN construction using Bundles and training.
After selecting top-$N$ promising Bundle candidates,
We search DNN models under resource and latency constraints.
For each Bundle, $K$ initial DNNs are generated and are progressively updated by adjusting the number of channels, pooling layer positions, etc., until the latency target is met.
Then, we perturb the initial DNNs by changing three variables:
the number of Bundle replications, 
down-sampling configurations between bundles, and channel expansion configuration. We adopted Stochastic Coordinate Descent (SCD) algorithm for perturbation, while other heuristic or evolutionary algorithms can be applied as well.
The goal of the search algorithm is to find the DNN architecture which meets the performance constraints with highest accuracy.

\begin{table*}[t]
\footnotesize
\centering
\caption{Performance Comparisons (FPGA and GPU competition data are obtained from \cite{xu2018dacsdc})}\label{tab:res-fpga-dnn}
\renewcommand{\arraystretch}{0.95}
\setlength{\tabcolsep}{4pt}
\begin{tabular}{|c | c | c c c c c c |}
\hline 
                         & {Model}  & {IoU} & {Latency (ms)}
                         & {FPS} & {Power} & {Energy}
                         & {Efficiency} \\ \hline
\multirow{6}{*}{Ours} 	 & DNN1 & 68.6\% & 80.0 (100 MHz) & 12.5 & 2.2W & 8.80 KJ & 0.18 J/pic \\
 						 & 					     &        & \textbf{57.4 (150 MHz)} & \textbf{17.4} & \textbf{2.5W} & \textbf{7.18 KJ} & \textbf{0.14 J/pic}  \\
  						 & DNN2 & 61.2\% & 62.6 (100 MHz) & 16.0 & 2.2W & 7.50 KJ & 0.15 J/pic \\
                         &                       &        & 44.1 (150 MHz) & 22.7 & 2.4W & 5.51 KJ & 0.11 J/pic  \\
  						 & DNN3 & 59.3\% & 47.8 (100 MHz) & 20.9 & 2.2W & 5.74 KJ & 0.11 J/pic  \\
                         &                       &        & \textbf{33.7 (150 MHz)} & \textbf{29.7} & \textbf{2.4W} & \textbf{4.04 KJ} & \textbf{0.08 J/pic}  \\

\hline
1st in FPGA 	 &  SSD & 62.4\% & 84.6 (150 MHz) & 11.96 & 4.2W & 17.56 KJ & 0.35 J/pic  \\
2nd in FPGA 	 & -- & 49.2\% & 38.5 (150 MHz) & 25.97 & 2.5W & 4.81 KJ & 0.10 J/pic  \\
3rd in FPGA 	 & -- & 57.3\% & 136.1 (150 MHz) & 7.35 & 2.6W & 17.69 KJ & 0.35 J/pic \\

\hline
1st in GPU 	 & Yolo & 69.8\% &  40.7 (854 MHz) & 24.55 & 12.6W & 25.66 KJ & 0.51 J/pic \\
2nd in GPU 	 & Tiny-Yolo & 69.1\% & 39.5 (854 MHz) & 25.3 & 13.3W  & 26.28 KJ & 0.53 J/pic \\
3rd in GPU 	 & Tiny-Yolo & 68.5\% & 42.3 (854 MHz) & 23.64 & 10.3W & 21.79 KJ & 0.44 J/pic \\
\hline 
\end{tabular}
\vspace{-12pt}
\end{table*}

\subsubsection{Evaluation results}

To evaluate the effectiveness of the co-design framework, we apply it on a low-power object detection competition~\cite{xu2018dacsdc}, and compare to the top-3 winners for both FPGA and GPU categories.
The results are shown in Table \ref{tab:res-fpga-dnn}.
We make comparisons in: (1) the Intersection over Union (IoU);
(2) the latency for processing one frame and the overall frame per second (FPS); 
(3) the board power;
(4) the energy consumption for all testing data;
and (5) the energy efficiency per frame (J/pic).
The results are collected from the board-level implementations on Pynq-Z1. 
The latency refers to the execution time for a single frame in millisecond, while FPS is measured using total run-time for the 50K images including image loading, preprocessing, and DNN inference.

Compared to the 1st-place winner of the FPGA category, we achieve 6.2\% higher IoU, 40\% lower power, and 2.5$\times$ better energy efficiency, which we attribute to the effectiveness of an automated co-search instead of manual designs.
Compared to GPU-based designs, our DNN1 model is more accurate than the 3rd-place design and only 1.2\% lower IoU than the 1st-place GPU design. 
Regarding the energy efficiency, ours is 3.6$\times$ better than the 1st-place GPU design with 40\% longer latency despite a nearly 6$\times$ slower clock frequency.

\subsection{EDD: Efficient Differential DNN Architecture Search}


\begin{figure}
    \centering
    \includegraphics[width=1\textwidth]{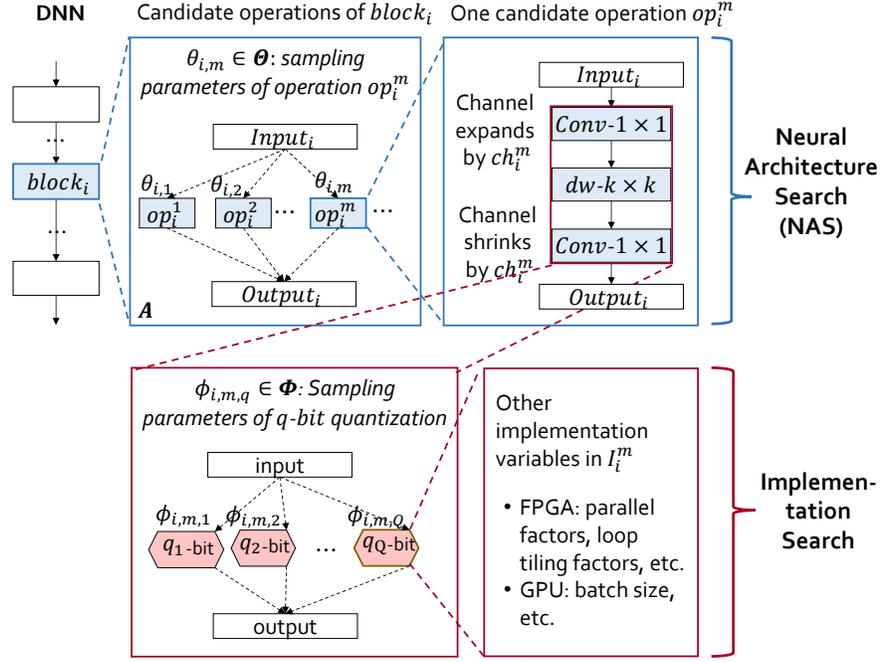}
    \caption{The overall architecture of EDD~\cite{li2020edd}.}
    \label{fig:edd-overall}
\end{figure}




On top of the FPGA/DNN co-design introduced in Sec.~\ref{sec:fpga-dnn-codesign}, we further develop co-design to a more generalized and unified approach, i.e., fully simultaneous neural architecture and implementation co-search, targeting arbitrary hardware platforms.
Neural architecture and implementation co-search (NAIS)~\cite{hao2019nais} is the first work that stylized design methodology targeting both FPGAs and GPUs, while EDD~\cite{li2020edd} is a fully simultaneous, efficient differentiable DNN architecture and implementation co-search methodology. 
The overall architecture of EDD is presented in Fig.~\ref{fig:edd-overall}.

\subsubsection{Fused co-design space}
The key technology is to \textit{fuse the design space of DNN architecture search and hardware implementation search}.
We collectively denote the variables used in DNN search and implementation search as $A$ and $I$, respectively, and the fused space of co-search is $\{A, I\}$.
To carry out both DNN architecture and hardware accelerator co-search in the fused DNN/accelerator space as described in E.~\ref{eq:hw-nas-obj}, we minimize the following loss function:
\begin{equation}
    min:\mathcal{L} = Acc_{loss}(A,I) \cdot Perf_{loss}(I) + \beta \cdot C^{RES(I) - RES_{ub}}
    \label{eq:co_search_obj_func}
    \vspace{-2pt}
\end{equation}
In the above equation, $Acc_{loss}$ is the DNN accuracy loss; $Perf_{loss}$ is the hardware performance loss such as end-to-end inference latency, throughput, energy, DNN model complexity, etc.; multiple performance metrics can be optimized simultaneously by defining a single weighted loss.
$RES$ is the resource utilization and $RES_{ub}$ is resource upper bound.
Apparently, $Acc_{loss}$ is a function of $A$ and $I$;
$Perf_{loss}$ and $RES$ are functions of $I$.
Resource upper-bound $RES_{ub}$ is expressed in an exponent term to introduce large penalty when being violated.
Worth noting, in the existing hardware-aware NAS approaches, only $A$ is searched while $I$ is \textit{fixed} during NAS.
In our proposed co-search formulation, $I$ is \textit{variable}, and $A$ and $I$ are \textit{fused as one design space $\{A, I\}$}.

\noindent
\textbf{NAS design space}.
In the search space, each DNN is composed of $N$ basic building blocks in a single-path fashion without branches~\cite{stamoulis2019single}.
Inside each block, there are $M$ candidate operations.
We adopt the most commonly used DNN blocks in NAS approaches, called MBConv~\cite{tan2019mnasnet}, which is composed of sequential layers of $conv$-$1\times1$, $dwconv$-$k\times k$ and $conv$-$1\times1$, where $k$ is the kernel size.
Between $conv$-$1\times1$ and $dwconv$-$k\times k$, the number of channels expands/shrinks by a ratio of $ch_i^m$ for operation $op_i^m$.
The output of each block is calculated based on the outputs of its $M$ candidate operations. Specifically, we adopt the Gumbel-Softmax function in \cite{wu2019fbnet}, where each operation $op_m^i$ will be sampled from a sampling parameter $\theta_{i, m}$ following Gumbel-Softmax distribution, which converts the discrete non-differentiable sampling to continuous differentiable sampling.
The sampled operations form a complete DNN, which can be evaluated for accuracy and implementation performance. 

\noindent
\textbf{Implementation search space}
We let each candidate operation $op_i^m$ has its own implementation variables, forming an implementation search space $I_i^m$.
The primary implementation variable is quantization $q$, i.e., data precision, since it has a large impact on DNN accuracy, implementation performance and hardware resource.
Rather than using a train-and-quantize approach, the quantization shall be searched together with DNN structure to provide implementation performance feedback.
Besides quantization, other implementation variables can also be integrated into the framework, such as accelerator parallelism, loop tiling factors, batch size, etc.

\subsubsection{Differentiable performance and resource formulation}

The key technology is how to formulate the loss function differentiable with respect to the search space $A$ and $I$.
Since NAS search space $A$ is discrete, differentiable formulation requires continuous relaxation. DARTS~\cite{liu2018darts} is the first work that uses softmax for relaxation, while FBNet uses Gumbel-softmax~\cite{jang2016categorical} by sampling from the discrete space.
Such relaxation has been demonstrated to be GPU-hours efficient with appealing model accuracy~\cite{liu2018darts, wu2019fbnet, li2021generic}. 
Motivated by FBNet, a similar technique using Gumbel-softmax can be applied to differentiable implementation $I$ to convert the discrete implementation search space into continuous. Therefore, by descending the loss function on validation set, $\{A,I\}$ can be learned simultaneously.

\subsubsection{State-of-the-art results}

\begin{table}[t]
    \centering
    \renewcommand{\arraystretch}{0.9}
    \small
    \caption{Comparisons with existing NAS solutions~\cite{li2020edd}.}
    \label{tab:nas_comp}
    \setlength{\tabcolsep}{2pt}
    \begin{tabular}{|c|c|c|c|c|}
    \hline
    & \multicolumn{2}{c|}{Test Error (\%)} & GPU Latency & FPGA Latency\\ 
    \hline
    & Top-1 & Top-5 & Titan RTX & ZCU102 \cite{CHaiDNN}\\
    \hline
    \multicolumn{5}{|l|}{\textbf{Baseline Models}}\\
    \hline
    GoogleNet  & 30.22 & 10.47 & 27.75 ms & 13.25 ms \\
    MobileNet-V2 \cite{sandler2018mobilenetv2}  & 28.1 & 9.7 & 17.87 ms & 10.85 ms\\
    ShuffleNet-V2 \cite{ma2018shufflenet} & 30.6 & 11.7 & 21.91 ms & NA \\
    ResNet18  & 30.2 & 10.9 & 9.71 ms & 10.15ms \\
    \hline
    \multicolumn{5}{|l|}{\textbf{Hardware-aware NAS Models}}\\
    \hline
    MNasNet-A1 \cite{tan2019mnasnet}  & 24.8 & 7.5 & 17.94 ms & 8.78 ms\\
    FBNet-C \cite{wu2019fbnet} & 24.9 & 7.6 & 22.54 ms & 12.21 ms\\
    Proxyless-cpu \cite{cai2018proxylessnas} & 24.7 & 7.6 & 21.34 ms & 10.81 ms \\
    Proxyless-Mobile \cite{cai2018proxylessnas} & 25.4 & 7.8 & 21.23 ms & 10.78 ms \\
    Proxyless-gpu \cite{cai2018proxylessnas} & 24.9 & 7.5 & 15.72 ms & 10.79 ms \\
    \hline
    \textbf{EDD-Net-1 } & 25.3 & 7.7 & \textbf{11.17 ms} & 11.15 ms \\
    \textbf{EDD-Net-2 } & 25.4 & 7.9 & 13.00 ms & \textbf{7.96 ms} \\

    \hline
    \end{tabular}
    \\
    \vspace{-4pt}
\end{table}

\begin{table}[t]

    \centering
    \renewcommand{\arraystretch}{0.9}
    \small
    \caption{EDD-Net-1 accuracy and latency on 1080 Ti~\cite{li2020edd}.}
    \label{tab:edd-net-1}
    \setlength{\tabcolsep}{2pt}
    \begin{tabular}{|c|c|c |c|}
    \hline
    & 32-bit Floating & 16-bit Floating & 8-bit Integer \\\hline
    Test Error & 25.5\% & 25.3\% & 26.4\% \\ \hline
    Latency & 2.83 ms &  2.29 ms & 1.74 ms \\

    \hline
    \end{tabular}
    \vspace{-4pt}
\end{table}

\begin{table}[t]

    \centering
    \renewcommand{\arraystretch}{0.9}
    \small
    \caption{Comparison of EDD-Net-3 with DNNBuilder~\cite{zhang2018dnnbuilder}}
    \label{tab:edd-net-3}
    \setlength{\tabcolsep}{2pt}
    \begin{tabular}{|c|c|c |c|}
    \hline
     & Top-1 Error (\%)  & Top-5 Error (\%) & Throughput (ZC706) \\\hline
    VGG16 & 29.5 & 10.0 & 27.7 fps \\\hline
    EDD-Net-3 & 25.6 & 7.7 & 40.2 fps\\

    \hline
    \end{tabular}
    \vspace{-4pt}
\end{table}

We demonstrate the results on a subset of ImageNet dataset randomly sampled from 100 classes and target three hardware architectures, each with a searched DNN model, called EDD-Net:
(1) low-latency oriented GPU (EDD-Net-1); 
(2) folded FPGA architecture (EDD-Net-2), where a single processing element (PE) will be reused by all layers;
(3) pipelined FPGA architecture (EDD-Net-3), where each layer has its own PE, and all PEs work simultaneously.
Each model is produced through EDD within a 12-hour search on a P100 GPU.

For GPU-targeted EDD-Net-1, the results are as shown in Table \ref{tab:nas_comp}, where the GPU latency is tested on Titan RTX.
It shows that EDD-Net-1 reaches similar or better accuracy comparing with the state-of-the-art DNN models and other NAS approaches while achieving the shortest inference latency.
Table \ref{tab:edd-net-1} shows the accuracy and latency tradeoff of different precisions of EDD-Net-1 on Nvidia 1080 Ti GPU.
For FPGA-targeted EDD-Net-2, the latency values are collected by running DNN models with CHaiDNN accelerators on ZCU102 FPGA as shown in Table \ref{tab:nas_comp}. It shows that EDD-Net-2 delivers the shortest latency on FPGA among all the DNNs. FPGA-targeted EDD-Net-3 is searched targeting a pipelined FPGA accelerator.
As shown in Table \ref{tab:edd-net-3}, EDD-Net-3 achieves higher throughput with a much higher accuracy comparing with the state-of-the-art.

\section{Conclusion}
\label{sec:conclusion}
Emerging DNN-based AI applications are challenging for embedded systems as these applications come with high computation and memory demands as well as diverse application-specific requirements, such as real-time responses, high-throughput performance, and reliable inference accuracy.
This chapter introduced a series of effective design methods to overcome these challenges to enable embedded AI solutions. These methods can be categorized into efficient machine learning algorithms, accelerator and compiler designs, and various co-design and optimization strategies.
We first proposed ELB-NN and VecQ to strengthen the AI model’s hardware efficiency by enabling extremely low bit-width quantization during model training and inference. 
Then, we proposed DNNBuilder and PyLog for customized hardware accelerator design and DNN workload mapping to such accelerators. 
At last, we introduced efficient co-design strategies, including FPGA/DNN co-design and EDD, when deploying AI workloads on embedded systems.

We believe embedded AI solutions will involve more effective and comprehensive design methods in the future, covering AI algorithms, customized accelerators, and co-design and co-optimization strategies between algorithm and hardware.
For example, our efficient AI algorithm designs, such as ELB-NN and VecQ, can adopt more advanced quantization schemes to minimize network compression loss. Future works will consider more diverse network architecture and layer-wise data distribution features. 
To facilitate a smoother accelerator design process, we will extend DNNBuilder and PyLog to create frameworks and tools for hardware design, synthesis, and workload compilation. Major directions include 1) heterogeneous computing support, which intends to enable system-level design and optimization for heterogeneous AI systems, and 2) dynamic computational graph scheduling, which enables the generation of runtime adaptive accelerators for future AI applications.
Our future works will also cover more advanced software/hardware co-design for emerging AI models running on heterogeneous systems, which contains a much larger design space and is thus more challenging. For example, multi-modal multi-task (MMMT) models \cite{hao2021software} and customized hardware designs \cite{talpes2020compute} working for autonomous driving have demonstrated the importance of heterogeneity in AI model and hardware designs. The co-design and co-optimization methods must be developed for such heterogeneous scenarios.

\bigskip
\small
\noindent \textbf{Acknowledgments} 
The works presented in this book chapter are mainly supported by the IBM-Illinois Center for Cognitive
Computing System Research (C3SR) – a research collaboration as part of IBM AI Horizons
Network, Semiconductor Research Corporation (SRC), the Campus for Research Excellence
and Technological Enterprise (CREATE) program in Singapore, AMD-Xilinx Center of Excellence, and a Google PhD Fellowship to Xiaofan Zhang. 
The authors also want to thank Chao Zhu, Cheng Gong, Chengyue Wang, Hyunmin Jeong, Jinjun Xiong, Junsong Wang, Kun Wu, Kyle Rupnow, Tao Li, Qiuwen Lou, Wen-mei Hwu, Xinheng Liu, Ye Lu, and Yonghua Lin for their valuable contributions.

\bibliography{main.bib}
\bibliographystyle{unsrt}

\end{document}